# Transfer Learning and Machine Learning for Training Five Year Survival Prognostic Models in Early Breast Cancer

Lisa Pilgram[1,2,3], Kai Yang[1,2], Ana-Alicia Beltran-Bless[4], Gregory R. Pond[5], Lisa Vandermeer[6], , John Hilton[7], Marie-France Savard[4], Andréanne Leblanc[8], Lois Sheperd[9], Bingshu E. Chen[9], John M.S. Bartlett[10], Karen J. Taylor[10], Jane Bayani[11,12], Sarah L. Barker[11], Melanie Spears[11,12], Cornelis J. H. van der Velde[13], Elma Meershoek-Klein Kranenbarg[13], Luc Dirix[14], Elizabeth Mallon[15], Annette Hasenburg[16], Christos Markopoulos[17], Lamin Juwara[1,2], Fida K. Dankar[2], Mark Clemons[4], Khaled El Emam[1,2]

[1]School of Epidemiology and Public Health, University of Ottawa, Ontario, Canada
[2]Children's Hospital of Eastern Ontario Research Institute, Ontario, Canada
[3]Department of Nephrology and Medical Intensive Care, Charité - Universitaetsmedizin Berlin, Berlin, Germany,
[4]Division of Medical Oncology, Department of Medicine, The University of Ottawa, Ontario, Ottawa, Canada
[5]Department of Oncology, McMaster University, Hamilton, Ontario, Canada
[6]Cancer Therapeutics Program, The Ottawa Hospital Research Institute, Ottawa, Ontario, Canada
[7]Ottawa Hospital Cancer Center, The Ottawa Hospital Research Institute, Ottawa, Ontario, Canada
[8]CHUM, Division of Medical Oncology and Hematology, Université de Montréal, Montreal, Quebec, Canada
[9]Canadian Canadian Cancer Trials Group, Queen's University, Kingston, Ontario, Canada
[10]Edinburgh Cancer Research, Institute of Genetics and Cancer, University of Edinburgh, Edinburgh, United Kingdom.
[11]Diagnostic Development, Ontario Institute for Cancer Research, Toronto, Ontario, Canada
[12]Department of Laboratory Medicine and Pathobiology, University of Toronto, Toronto, Ontario, Canada.
[13]Department of Surgery, Leiden University Medical Center, Leiden, the Netherlands.
[14] St. Augustinus Hospital, Antwerp, Belgium.
[15]Department of Pathology, Glasgow, United Kingdom
[16]Department of Gynecology and Obstetrics, University Center Mainz, Mainz, Germany.
[17]National and Kapodistrian University of Athens, Medical School, Athens, Greece.

Corresponding Author:
Khaled El Emam
CHEO Research Institute
401 Smyth Road
Ottawa, Ontario K1H 8L1
Canada
kelemam@ehealthinformation.ca

## Abstract

**Background:** Prognostic information is essential for decision-making in breast cancer management. In recent years, trials have predominantly focused on genomic prognostication tools, even though clinicopathological prognostication is less costly and more widely accessible. Advances in machine learning (ML), transfer learning and ensemble integration now offer opportunities to build robust prognostication frameworks, particularly in contexts where missingness and model assumptions vary across cohorts.

**Objectives:** This study evaluates this potential to improve survival prognostication in breast cancer, more precisely we compare de-novo ML, transfer learning from a pre-trained prognostic tool and ensemble integration.

**Methods:** Data from the MA.27 trial (NCT00066573) was used for model training, with external validation on data from the TEAM trial (NCT00279448, NCT00032136) and a SEER cohort. Transfer learning was applied by fine-tuning the pre-trained prognostic tool *PREDICT v3*, de-novo ML included Random Survival Forests (RSF) and Extreme Gradient Boosting (XGB), and ensemble integration was realized through a weighted sum of model predictions. The Integrated Calibration Index (ICI) was used as optimization goal during training; internal and external validation was assessed in terms of calibration (ICI) and discrimination (area under the receiver operating characteristic curve, AUC). Shapley Additive Explanations (SHAP) were used to explain model predictions.

**Results:** Transfer learning, de-novo RSF, and ensemble integration relevantly improved calibration in MA.27 over the pre-trained model (ICI reduced from 0.042 in *PREDICT v3* to ≤0.007) while discrimination remained comparable (AUC increased from 0.738 in *PREDICT v3* to 0.744-0.799). Invalid *PREDICT v3* predictions were observed in 23.8-25.8% of MA.27 individuals due to missing information. In contrast, ML models and ensemble integration could predict survival regardless of missing information. Across all models, patient age, nodal status, pathological grading and tumor size had consistently highest SHAP values, indicating their importance for survival prognostication. External validation in SEER, but not in TEAM, confirmed the benefits of transfer learning, RSF and ensemble integration in terms of calibration.

**Conclusions:** This study demonstrates that transfer learning, de-novo RSF, and ensemble integration can improve prognostication in situations where relevant information for *PREDICT v3* is lacking or where a dataset shift is likely. Ultimately, better survival estimation can provide meaningful guidance in breast cancer treatment, supporting a more targeted, cost-effective, and personalized approach to breast cancer care.

# 1. Introduction

Breast cancer is among the most common types of cancer worldwide. In 2022, there were 2.3 million women diagnosed with breast cancer globally [1] typically in a non-metastatic stage at diagnosis [2]. Such early diagnosis allows for a broad range of treatment options, including surgery, radiation therapy, endocrine therapy, chemotherapy and targeted systemic therapies. Estimating survival probabilities can support informed decision-making, especially in scenarios where multiple treatment options are available. This makes both, prognostic information (a patient's survival) and predictive information (a patient's benefit from treatment) [3], highly valuable to guide patient management.

A large variety of clinicopathological and genomic risk assessment tools have been proposed to assist in clinical decision-making [3,4]. Other tools, for example, RSClin have been developed to improve prognosis by incorporating both, clinicopathological and genomic information [5].

Trials like MINDACT [6] confirmed the importance but also the challenge of breast cancer risk assessment tools. It demonstrated that patients classified as low risk by genomic prognostication but as high risk by clinicopathological prognostication had favorable survival outcomes even without chemotherapy. Conversely, patients classified as high risk by genomic prognostication and low risk by clinicopathological prognostication had favorable survival outcomes as well whether or not they had received chemotherapy. Similarly, the recently published ASTER 70s trial shows that women aged 70 years and above who had a high risk according to genomic prognostication did not benefit from the addition of adjuvant chemotherapy to endocrine therapy in terms of survival [7]. Both trials underscore the limitations of current genomic prognostication tools in predicting treatment benefit and reinforce the critical distinction between prognostic and predictive information [3].

Despite these findings, genomic testing and prognostication has become routine even for clinicopathologically low-risk patients, diverging from clinical guidelines recommendations and thereby potentially resulting in over-treatment [4-6]. Also, genomic testing can come with delayed treatment decisions and considerable costs, limiting their accessibility in resource-constrained healthcare settings [8].

Importantly, recent trials have focused primarily on genomic prognostication tools. There has been relatively less interest in the refinement of clinicopathological prognostication [9], even though such solutions are inexpensive, easily accessible, and can support clinical decision-making. Recently, machine learning (ML) and deep learning approaches have been explored to enhance performance in survival prognostication with promising results but limited applicability in practice [10–12] (see **Appendix A: Supplemental Background – Machine Learning in Survival Analysis**).

A compelling opportunity also lies in leveraging validated pre-trained models like *PREDICT v3* [13] and adapting them to new datasets using parameter-based transfer learning (i.e., fine-tuning) [14]. In transfer learning, the idea is that a new task can be more effectively learned by transferring knowledge from a related task that has already been learned. There are multiple transfer learning approaches (see **Appendix A: Supplemental Background – Transfer Learning in Survival Analysis**). Among them, parameter-based transfer learning is most useful, as it does not require access to the original training datasets but instead fine-tunes the parameters of the pre-trained model to the new data [14]. While such transfer learning is not yet widely adopted in survival analysis, there are promising results from lung cancer and pancreatic adenocarcinoma survival prognostication [15,16].

Complementary to de-novo ML and transfer learning, ensemble integration offers a way to account for model-specific strengths and limitations by combining multiple models such as fine-tuned and de-novo trained ones [17]. This can ultimately provide a robust prognostication framework, particularly in cases where missingness and model assumptions vary across cohorts.

This study aims at better understanding the benefits of transfer learning from pre-trained models, de-novo ML, and ensemble integration in survival prognostication for patients with breast cancer. More precisely, using the MA.27 study population as a test case [18], we investigated the following research questions:

1. Parameter-based transfer learning (i.e., fine-tuning): Can fine-tuning the pre-trained prognostic tool *PREDICT v3* to the MA.27 dataset improve survival prediction performance compared to the pre-trained model alone?
2. De-novo ML: How do state-of-the-art ML models trained directly on the MA.27 dataset compare against the (fine-tuned) pre-trained model *PREDICT v3*?
3. Ensemble integration: Does an ensemble of fine-tuned pre-trained models and de-novo ML models add benefit compared to either approach alone?
4. Generalizability: Do the potential benefits from fine-tuning (i.e. question 1), de-novo ML (i.e., question 2), and ensemble integration (i.e., question 3) still hold in comparable external cohorts?

Ultimately, better survival estimation can provide meaningful guidance in breast cancer treatment, supporting a more targeted, cost-effective, and personalized approach to breast cancer care.

## 2. Methods

### 2.1. MA.27 Study Population

MA.27 was a phase 3 clinical trial conducted by the Canadian Cancer Trials Group (CCTG) [18] of 7,576 postmenopausal women with early-stage hormone receptor-positive breast cancer between 2003 and 2008, which compared two aromatase inhibitors, exemestane and anastrozole, as adjuvant endocrine therapy.

We limited the cohort to patients who survived the day of enrollment (i.e., time-to-event > 0) for our study to avoid biasing survival estimates with events unrelated to breast cancer. This sub cohort consisted of 7,563 patients. For this sub cohort, variables were selected from MA.27 that overlapped with the information required for *PREDICT v3* (see **Table 1** and details in **Appendix A: Supplemental Methods – Pre-trained Model and Transfer Learning**). This ensured alignment with variables considered clinically relevant and broadly available in the context of breast cancer prognostication. The outcome (i.e., event) was defined as breast cancer-related death within a 5-year observation interval, and time-to-event or follow-up-time was considered in the survival analyses.

| Variable | Explanation |
|---|---|
| Age | Age in years at randomisation |
| Positive nodes | The number of positive lymph nodes |
| Tumor laterality | Side of tumor manifestation: right handed, left handed or bilateral |
| Estrogen receptor status | Positive or negative estrogen receptor status |
| Progesterone receptor status | Positive or negative progesterone receptor status |
| Tumor size | The maximum size of the tumor in mm |
| Tumor grade | The pathological grading of the tumor from 1 (well-differentiated) to 3 (poorly differentiated) |
| Radiotherapy | Whether or not radiotherapy was received |
| Chemotherapy | Whether or not adjuvant chemotherapy was received |
| Trastuzumab therapy | Whether or not trastuzumab was received |

**Table 1. Variables Selected for 5-Year Survival Prediction**. These variables represent the overlap between variables in MA.27 and those required for *PREDICT v3*. The variables required for *PREDICT v3* but not directly available were year of diagnosis, smoking status, HER2 status, Ki-67 status, mode of detection, micrometastases in case of one positive node, mean heart dose in case of radiotherapy, the type of chemotherapy in case of chemotherapy and bisphosphonate use. Some of them were mandatory for survival prognostication, so that assumptions were made based on standard of care at that time. Details are provided in **Appendix A: Supplemental Methods – Pre-trained Model and Transfer Learning**.

## 2.2. Data Management

MA.27 was highly imbalanced in terms of its outcome meaning that there were only 187 (2.5%) recorded disease-related deaths across a median follow-up of 4.1 years. The Random Over-Sampling Examples (ROSE) technique can provide a more balanced and representative distribution of outcome to help with model training [19] (see **Appendix A: Supplemental Methods – Outcome Re-Balancing**). We tested the effect of ROSE in model training but could not detect a beneficial effect so it was not considered in the main analyses (see **Appendix A: Supplemental Results – Re-Balancing For Model Training**).

MA.27 further presented with missingness in some variables. The mechanism of missingness was, however, unclear. A missing progesterone receptor status could be, for example, not missing at random if it reflected ambiguity in the pathological assessment, suggesting a potentially biologically meaningful pattern of missingness. It could, however, also be missing at random if values were omitted due to documentation errors or data entry inconsistencies. To avoid the introduction of bias, we therefore refrained from using imputation but leveraged the abilities of tree-based ML model to internally handle missing data via surrogate splits or default directions as they can handle mixed types of missingness patterns [20].

As mentioned in **Table 1** and detailed in **Appendix A: Supplemental Methods – Pre-trained Model and Transfer Learning**, the overlap between the variables in MA.27 and those required for *PREDICT v3* was not complete, so that some variables were constructed based on the trial's metadata and relevant background knowledge. For example, HER2 status was inferred from trastuzumab use, and endocrine therapy from the inclusion criteria of MA.27. For other variables, no reliable approximation was possible, so that a fraction of patients for whom *PREDICT v3* could not estimate survival remained.

## 2.3. Survival Models

*PREDICT* was originally fitted on 5,232 breast cancer cases from the UK East Anglia Cancer Registration and Information Centre diagnosed in 1999-2003 [13], updated recently to *PREDICT v3* with 38,909 patients diagnosed between 2000 and 2017 [21] and validated on several cohorts around the world [22,23]. MA.27, however, differs in certain aspects from these training and validation cohorts: MA.27 was collected in 2003, involved approximately 5 years of follow-up, included only postmenopausal hormone positive patients, and lacked some of the information required in *PREDICT v3*. This makes de-novo ML and transfer learning particularly useful. **Appendix A: Supplemental Background** gives supplemental background on commonly applied survival models including *PREDICT*, de-novo ML and transfer learning.

Given the structure of the MA.27 dataset with predominantly categorical variables, high censoring and limited sample size, we opted to use tree-based methods for ML survival modeling. Such models are good in handling categorical data, can internally account for mixed missingness types, and offer robust performance without the data demands or complexity of deep learning approaches.

More precisely, this study included the following models:

- *PREDICT v3*: The pretrained survival model.
- *f-PREDICT v3*: The pretrained survival model fine-tuned to MA.27 (i.e., transfer learning)
- Random Survival Forests (RSF) [24]: An ensemble method tailored for survival analysis.
- Extreme Gradient Boosting (XGB) [25]: A gradient boosting framework with survival-specific loss function.
- Ensemble [17]: An ensemble integrating *f-PREDICT v3*, RSF and XGB whereby final predictions are obtained as a weighted sum of the individual model predictions

Details on implementation are provided in **Appendix A: Supplemental Methods – Machine Learning Survival Models**.

## 2.4. Performance Measurement and Model Explainability

The primary goal of this study was to predict 5-year breast cancer survival, as this represents an early and clinically relevant milestone to guide decision-making. Accordingly, performance metrics focused on this time-point.

In internal and external validation, we measured performance for 5-year-survival prediction by Area Under the Receiver Operating Characteristic (ROC) Curve (AUC) and Integrated Calibration Index (ICI) and provide calibration plots and ROC curves for that time point. AUC reflects a model's ability to distinguish between individuals who survive and those who do not (i.e., discrimination). ICI is the average absolute difference between predicted survival probabilities and observed survival outcome, estimated from a smoothed calibration curve over the entire range of predictions [26]. All performance measurements accounted for the time-to-event nature of the data in order to obtain robust measures in the context of heavy censoring. For discrimination, inverse probability of censoring weighting (IPCW) was leveraged to adjust for right-censoring whereby the Kaplan-Meier estimator was used to model the censoring distribution [27]. It was implemented via the R package timeROC [28]. For calibration, the observed outcome was modeled as proposed in [26].

Our interpretation of AUC are informed by a literature review on interpreting AUC in healthcare [29] whereby changes in label were typically triggered by a change in AUC of at least 0.1. The interpretation of ICI is more challenging as no uniform guidance exists. However, in experiments by Austin et al. [26,30] correctly specified models yielded values below 0.0125 while incorrectly specified ones had higher values.

Shapley Additive Explanations (SHAP) is a post-hoc method to explain model predictions based on game theory [31]. For each individual prediction, a Shapley value can be assigned to each variable that reflects the contribution of that very variable to the prediction. This value is estimated by a Monte Carlo approximation strategy as suggested in [31]. A model-agnostic SHAP approach was leveraged to ensure consistent interpretation across the five different models. It was implemented via the R package iml [32]. SHAP is presented for all individuals from MA.27 in a summary plot, implemented via shapviz [33].

## 2.5. Model Training, Testing and Internal Validation

MA.27 was randomly split into three subsets: 60% for training (data A), 20% for testing (data B), and 20% for final validation (data C).

As shown in **Figure 1**, data A was used to train ML models and to fine-tune *PREDICT v3* (i.e., transfer learning). The fine-tuned model is referred to as *f-PREDICT v3*. Fine-tuning was performed by adjusting the 26 parameters of *PREDICT v3* through a local optimization approach to MA.27 (see **Appendix A: Supplemental Methods – Pre-Trained Model and Transfer Learning**).

Data B was used to find the best hyperparameters for the ML models and to determine the weights for integrating RSF, XGB and *f-PREDICT v3* into an ensemble. Details on hyperparameter and ensemble integration are provided in **Appendix A: Supplemental Methods – Machine Learning Survival Models**.

ML training with the best hyperparameters and fine-tuning of *PREDICT v3* were repeated on the combined data A and B prior to the internal evaluation (i.e., validation) on the hold-out validation set (data C). After internal evaluation, the final weights for ensemble integration were searched based on the combined data A and B, and ML training and fine-tuning of *PREDICT v3* were conducted on the entire MA.27 dataset.

During training, the ICI for 5-years-survival prediction was used as the optimization goal. Calibration has clinically meaningful implications in prognostication tools as probability estimates typically guide decision making. Discrimination (i.e., AUC), in contrast, focuses on ranking which may be more relevant for diagnostic tools. Other metrics, such as the Mean Absolute Error (MAE) assess the accuracy of the predicted survival times, which, again, is different from our scenario where survival probabilities at certain time points are relevant for decision making [34]. An AUC-based training approach was also conducted and is presented in **Appendix A: Supplemental Results – AUC Optimization for Model Training**. Both, calibration (i.e., ICI) and discrimination (AUC), were, however, part of evaluation.

To account for the variability in modeling across different splits, all training, testing, and validation steps were repeated across 10 independent runs. The ML parameters were chosen by majority vote (i.e., the parameter that was most often chosen across the 10 independent runs), the fine-tuned parameters for *PREDICT v3* and the ensemble weights by averaging. We indicate the median and interquartile range (IQR) for the internal evaluation across the 10 independent runs.

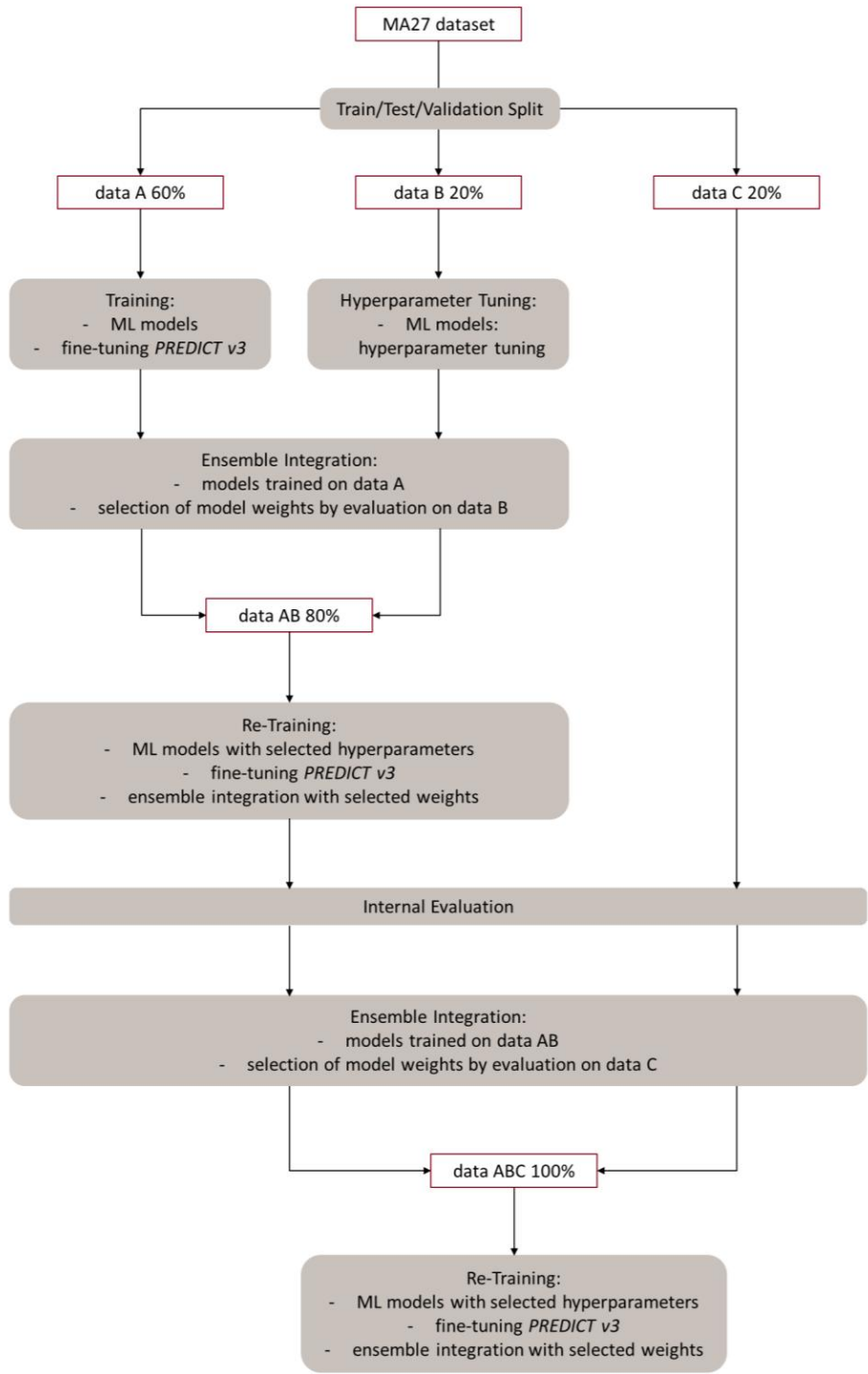

**Figure 1. Training, Testing and Validation for MA.27**. The MA.27 dataset was split into a training dataset (60%) used to train the ML models and to fine-tune *PREDICT v3*, a testing dataset (20%) used to find hyperparameters for the ML models and weights for the ensemble, and a validation dataset (20%) for internal evaluation. ML training and fine-tuning are repeated after internal validation based on the entire dataset. Ensemble integration on the combined data A and data B.

## 2.6. External Validation

The final models were externally validated against a similar cohort from the US Surveillance, Epidemiology, and End Results (SEER) program [35] and a similar clinical trial cohort (Tamoxifen Exemestane Adjuvant Multinational, TEAM) with patients from Belgium, France, Germany, Greece, Japan, the Netherlands, the UK and Ireland, and the USA [36].

In SEER [35], the cohort was selected to match the relevant eligibility criteria of MA.27, namely hormone receptor positive postmenopausal women and diagnosed in 2003. It was also adjusted to meet the requirements of *PREDICT v3* that should not be used in DCIS/LCIS only or in women with metastatic disease. This cohort comprised 27,064 individuals.

We used data from a TEAM sub-study [9]. TEAM [36] represented a similar cohort as MA.27 so that the cohort was not further sub-selected. In total, 3,825 individuals from TEAM were included in our analyses.

Information about missingness and variable mapping to *PREDICT v3* is provided in **Appendix A: Supplemental Methods – Pre-Trained Model and Transfer Learning**. The external validation included calibration (i.e., ICI) and discrimination (i.e., AUC) as detailed above. 95% CI were further derived from bootstrapping by taking the 2.5% and 97.5% percentile of the bootstrap distribution.

# 3. Results

## 3.1. Description of MA.27 Study Population

Our MA.27 dataset included 7,563 postmenopausal women diagnosed with breast cancer. **Table 2** provides an overview of the patients' characteristics.

| Variable | Median (IQR) / Number of Patients (Percentage) |
|---|---:|
| Age | 64.2 (IQR 58.2-71.2) |
| Nodal stage | |
| N0 | 5,360 (71.9%) |
| N1 | 1,615 (21.7%) |
| N2 | 357 (4.8%) |
| N3 | 124 (1.6%) |
| Tumor laterality | |
| Left handed | 3,785 (50.1%) |
| Right handed | 3,663 (48.4%) |
| Bilateral | 115 (1.5%) |
| Hormone receptor status | |
| Estrogen receptor status positive | 7,513 (99.3%) |
| Progesterone receptor status positive | 6,079 (82.0%) |
| Tumor size in cm | 1.5 (IQR 1.0-2.0) |
| Tumor grade | |
| G1 | 1,892 (32.0%) |
| G2 | 2,982 (50.5%) |
| G3 | 1,036 (17.5%) |
| Therapy | |
| Radiotherapy | 5,370 (71.1%) |
| Chemotherapy | 2,326 (30.8%) |
| Trastuzumab therapy | 67 (3.5%) |
| Event | 187 (2.5%) |

**Table 2. Characteristics of MA.27 Study Population.** Percentages are calculated excluding missing values. The event was defined as breast cancer-related death within a 5-year observation interval. Details on missing values are provided in **Appendix A: Supplemental Methods – Pre-Trained Model and Transfer Learning**.

## 3.2. Performance Across Transfer Learning, De Novo ML, and Ensemble Integration

We evaluated the prognostic performance of three potential improvement strategies, transfer learning or fine-tuning (*f-PREDICT v3*), de-novo ML (RSF, XGB) and ensemble integration, and compared them to the pre-trained model *PREDICT v3*. This evaluation was done to identify the most effective strategy when conducting prognostication in a new cohort where the lack of certain information and distribution shifts can considerably compromise the performance of pre-trained models.

We also tested re-balancing as a strategy to improve the training process but could not detect a beneficial effect (see **Appendix A: Supplemental Results – Re-Balancing For Model Training**). In fact, model performance in terms of calibration decreased by large (more than 10 times increase in ICI median up to 0.554) while AUC remained mainly unchanged (changes in AUC median less than 0.1).

In the following, we present the evaluation of the pre-trained model itself as well as transfer learning, de-novo ML and ensemble integration when training was optimized for ICI without re-balancing.

**Table 3** gives the summary of calibration (i.e., ICI) and discriminative performance (AUC) in the form of the median across the 10 independent runs as explained in the methods section. In terms of calibration, fine-tuning (i.e., *f-PREDICT v3*) and RSF presented with best results in the internal evaluation (ICI 0.005 and 0.003, respectively). Performance in terms of discrimination ranged from an AUC of 0.738 (*PREDICT v3*) to an AUC of 0.799 (*f-PREDICT v3*). Ensemble integration of *f-PREDICT*, RSF and XGB yielded results comparable to the best stand-alone models (ICI 0.007 and AUC 0.746).

| Model | Calibration (ICI) | | Discrimination (AUC) | |
|---|---|---|---|---|
| | Median | IQR | Median | IQR |
| *PREDICT v3* | 0.042 | 0.039-0.047 | 0.738 | 0.719-0.719 |
| *f-PREDICT v3* | 0.005 | 0.004-0.010 | 0.799 | 0.789-0.818 |
| RSF | 0.003 | 0.002-0.008 | 0.744 | 0.731-0.760 |
| XGB | 0.040 | 0.038-0.043 | 0.783 | 0.764-0.810 |
| Ensemble | 0.007 | 0.003-0.009 | 0.746 | 0.733-0.766 |

**Table 3. Calibration and Discrimination for 5-Year Survival**. Values were calculated on the validation dataset. The median and IQR across 10 seed settings is indicated for all survival models. Training was done on the non-imputed and non-rebalanced dataset and optimized for ICI.

Consequently, all three improvement strategies added value compared to the pre-trained model *PREDICT v3* in MA.27. Calibration plots are illustrated in **Figure 2** for all survival models, and ROC curves in **Figure 3,** confirming the summary results.

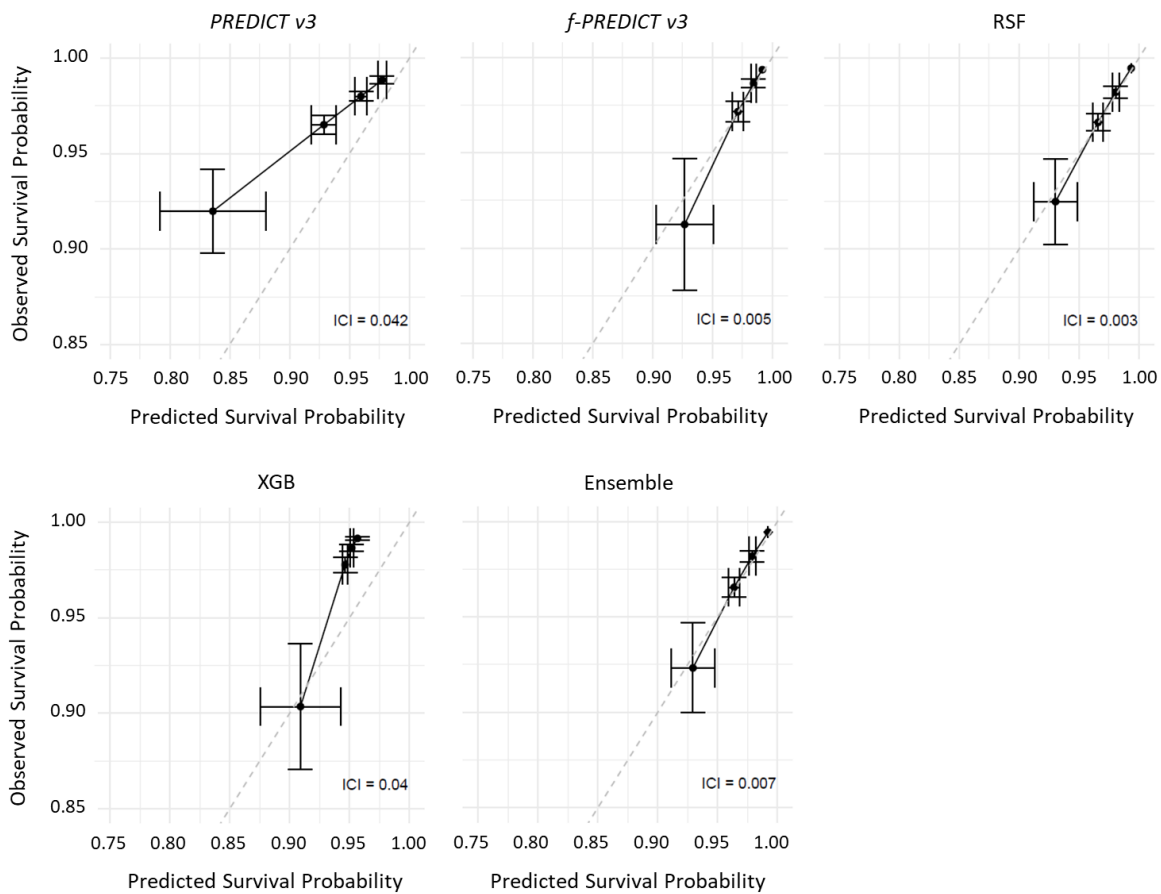

**Figure 2. Calibration Plots for 5-Year Survival.** Calibration plots are illustrated for the baseline model *PREDICT v3* as well as the four models enhanced by transfer learning (i.e., *f-PREDICT v3*), ML (RSF and XGB) and ensemble integration. The diagonal dashed line indicates perfect calibration. Data was pooled across the 10 independent runs. Observed probabilities were smoothed using a hazard regression-based method [26]. Observations were then divided into four quartiles based on their predicted probabilities. Both predicted and observed probabilities were trimmed to exclude extreme values beyond the 10th to 90th percentile, and mean predicted and observed survival probabilities were calculated in each quartile. The horizontal and vertical error bars reflect the standard deviations of these quartile-wise means. The median ICI across the 10 independent runs is given for each model.

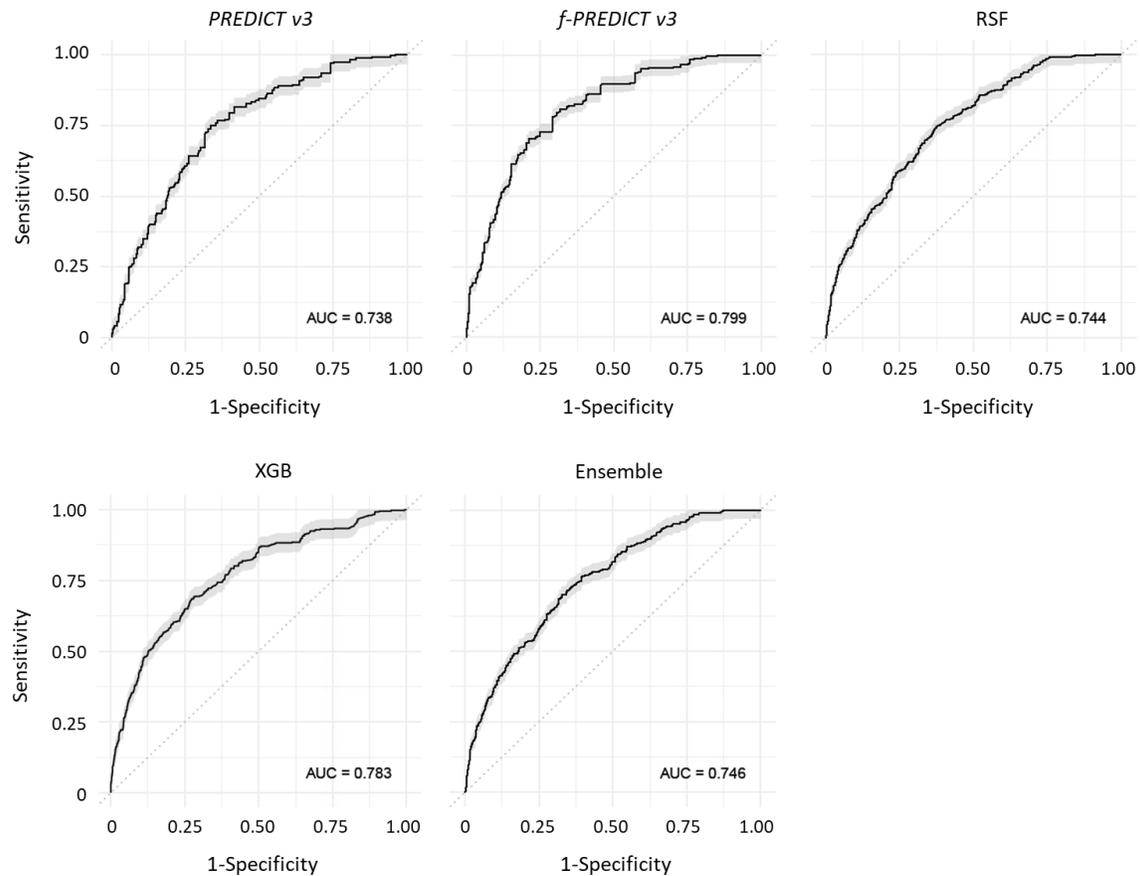

**Figure 3. Receiver Operating Characteristics (ROC) Curves for 5-Year Survival.** ROC curves are illustrated for the baseline model *PREDICT v3* as well as the four models enhanced by transfer learning (i.e., *f-PREDICT v3*), ML (RSF and XGB) and ensemble integration. Data was pooled across the 10 independent runs to plot ROC curves. The diagonal dashed grey line indicates discrimination of a random guess. The median AUC across the 10 independent runs is given for each model.

MA.27 had incomplete records with respect to some variables that were required for *PREDICT v3* for valid survival estimation. Consequently, *PREDICT v3* and *f-PREDICT v3* could not give survival estimates for a subset of the entire dataset. Such invalid predictions occurred in patients where relevant information was missing (see methods). The number of invalid predictions in *PREDICT v3* and *f-PREDICT v3* ranged from 361 (23.8%) to 391 (25.8%) across the 10 independent runs. In contrast, RSF, XGB and the ensemble integration could predict survival for all individuals, independent of missing information.

We provide performance evaluation stratified by whether or not *PREDICT v3* and *f-PREDICT v3* returned valid predictions in **Table 4**. In general, models performed worse in patients that were lacking relevant information, but these differences were very small and RSF and the ensemble integration still presented with good calibration (ICI 0.014 and 0.015 respectively) in this subset.

| Model | Calibration (ICI) | | Discrimination (AUC) | |
|---|---|---|---|---|
| | Median | IQR | Median | IQR |
| Subset with valid predictions in *PREDICT v3* | | | | |
| *PREDICT v3* | 0.042 | 0.039-0.047 | 0.738 | 0.719-0.719 |
| *f-PREDICT v3* | 0.005 | 0.004-0.010 | 0.799 | 0.789-0.818 |
| RSF | 0.006 | 0.004-0.009 | 0.763 | 0.748-0.781 |
| XGB | 0.043 | 0.040-0.044 | 0.802 | 0.786-0.812 |
| Ensemble | 0.006 | 0.005-0.014 | 0.775 | 0.760-0.796 |
| Subset with invalid predictions in *PREDICT v3* | | | | |
| RSF | 0.014 | 0.013-0.015 | 0.726 | 0.713-0.713 |
| XGB | 0.051 | 0.039-0.057 | 0.745 | 0.628-0.825 |
| Ensemble | 0.015 | 0.015-0.021 | 0.691 | 0.668-0.739 |

**Table 4. Calibration and Discrimination in Subsets With and Without Relevant Missing Information**. Values were calculated on the validation dataset stratified by whether or not *PREDICT v3* could provide survival estimates (i.e., valid versus invalid predictions). Invalid predictions occurred in patients where relevant information was missing. The median and IQR across 10 seed settings is indicated for all survival models. Training was done on the non-imputed and non-rebalanced dataset and optimized for ICI.

### 3.3. Model Explainability

Results from the SHAP analysis are illustrated in **Figure 4**. While the exact top 3 variables varied slightly between models, patient age, nodal status, pathological grading and tumor size were consistently among them. In contrast, treatment information such as chemotherapy, radiotherapy or trastuzumab were typically ranked less important across all models.

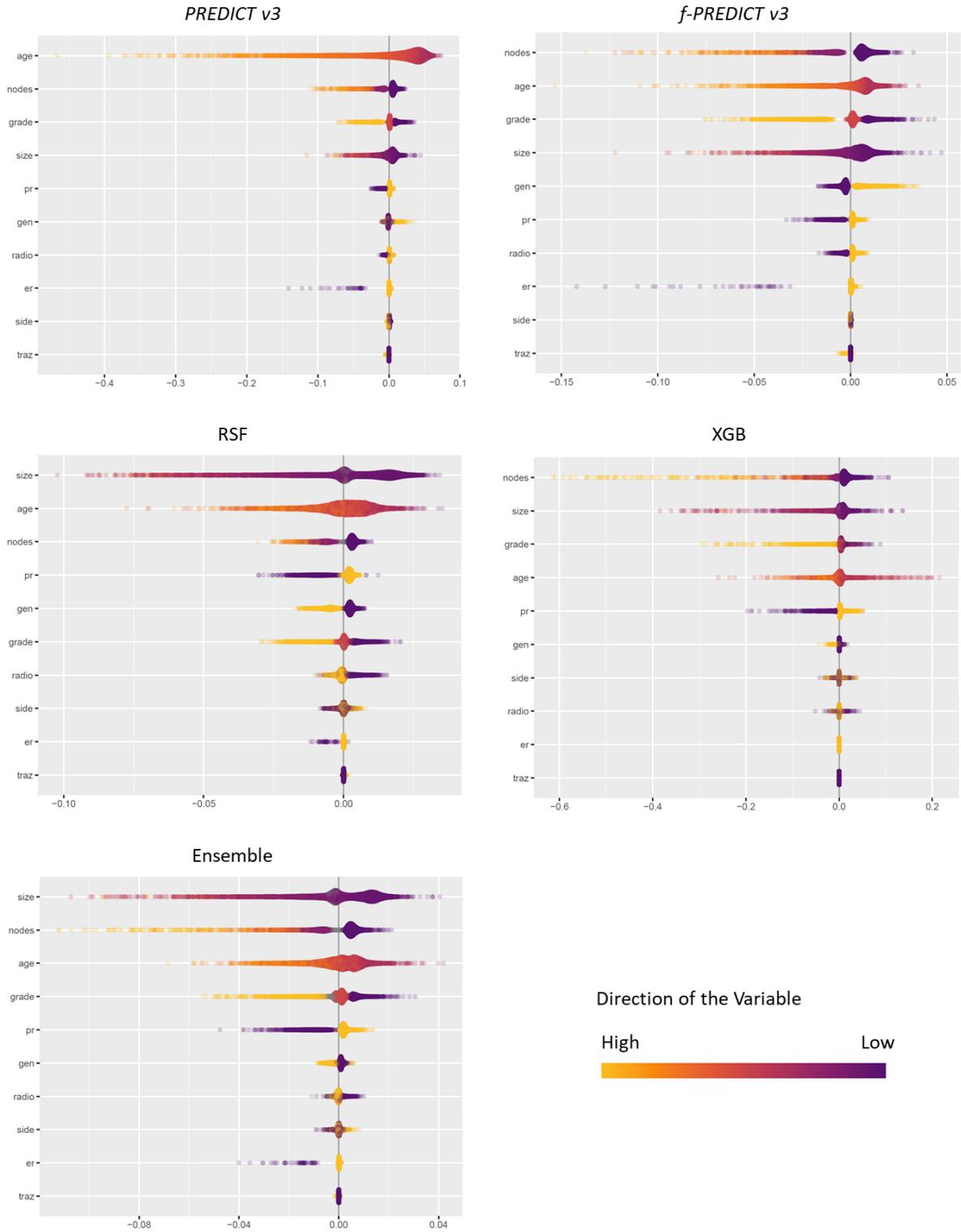

**Figure 4. SHAP analysis for MA.27.** Shapley values were calculated for all records in MA.27. The variables are ranked based on their mean absolute SHAP values across all observations, reflecting the importance of the variable for the entire MA.27 dataset. The x-axis indicates the Shapley value, where its size reflects the strength of a variable's contribution to the prediction, and the sign indicates the direction of this contribution (i.e., positive Shapley values increase the survival probability). The direction of the variable is color-coded. Er: estrogen receptor status; gen: chemotherapy; grade: pathological grading; nodes: number of positive nodes; pr: progesterone receptor status; radio: radiotherapy; side: tumor laterality; size: size of tumor; traz: trastuzumab therapy.

### 3.4. External Validation on SEER and TEAM

To assess the extent to which the MA.27 model optimization was generalizable, we externally validated all models on data from the US SEER program and on the clinical trial dataset TEAM.

The SEER cohort was selected in line with the MA.27 eligibility criteria (i.e., postmenopausal women with non-metastatic hormone-receptor positive breast cancer diagnosed in 2003). Results are shown in **Table 5** and confirmed the benefit of transfer learning, de novo ML and ensemble integration over the pre-trained model. Calibration was worse than in the internal evaluation with an ICI of 0.010 (*f-PREDICT v3*), of 0.020 (RSF) and 0.018 (ensemble); discriminative performance comparable to the one in the internal evaluation.

| Model | Calibration | | Discrimination | |
|---|---|---|---|---|
| | ICI | 95% CI | AUC | 95% CI |
| *PREDICT v3* | 0.039 | 0.037-0.198 | 0.765 | 0.750-0.780 |
| *f-PREDICT v3* | 0.010 | 0.009-0.173 | 0.825 | 0.811-0.838 |
| RSF | 0.020 | 0.019-0.046 | 0.753 | 0.741-0.765 |
| XGB | 0.037 | 0.034-0.093 | 0.759 | 0.747-0.771 |
| Ensemble | 0.018 | 0.016-0.078 | 0.792 | 0.779-0.802 |

**Table 5. Calibration and Discrimination for SEER Data**. Values were calculated on the external validation data. The value and its 95% CI as derived from bootstrapping is indicated for all survival models. Training was done on the non-imputed and non-rebalanced MA.27 dataset and optimized for ICI.

For TEAM, respective results are given in **Table 6**. Calibration plots and ROC curves are given in **Appendix A: Supplemental Results – External Evaluation Plots**. In contrast to SEER, the MA.27 optimized models performed worse on the TEAM dataset with changes in AUC up to 0.122 (XGB from 0.783 in internal evaluation to 0.661 in external validation via TEAM) and changes in ICI up to 0.074 (*f-PREDICT* from 0.005 in internal evaluation to 0.079 in the external validation via TEAM).

| Model | Calibration | | Discrimination | |
| --- | --- | --- | --- | --- |
| | ICI | 95% CI | AUC | 95% CI |
| *PREDICT v3* | 0.034 | 0.028-0.058 | 0.701 | 0.672-0.732 |
| *f-PREDICT v3* | 0.079 | 0.072-0.089 | 0.707 | 0.677-0.736 |
| RSF | 0.073 | 0.071-0.076 | 0.648 | 0.623-0.673 |
| XGB | 0.061 | 0.051-0.091 | 0.661 | 0.635-0.687 |
| Ensemble | 0.073 | 0.071-0.076 | 0.678 | 0.651-0.703 |

**Table 6. Calibration and Discrimination for TEAM Data**. Values were calculated on the external validation data. The value and its 95% CI as derived from bootstrapping is indicated for all survival models. Training was done on the non-imputed and non-rebalanced MA.27 dataset and optimized for ICI.

## 4. Discussion

### 4.1. Summary and Comparison to Literature

This study investigated how innovative learning approaches, including pre-trained models combined with transfer learning, de novo ML (RSF, XGB) and ensemble integration, can be leveraged to enhance the performance of prognostication tools for breast cancer. Using the MA.27 dataset, we addressed four questions: First, can fine-tuning the pre-trained prognostic tool PREDICT v3 to the MA.27 dataset improve survival prediction performance compared to the pre-trained model alone? Second, how do state-of-the-art ML models trained directly on the MA.27 dataset compare against the (fine-tuned) pre-trained model *PREDICT v3*? Third, does integrating fine-tuned pre-trained models and de-novo ML models using ensemble learning add benefit compared to either approach alone? And fourth, do the potential benefits from fine-tuning, de-novo ML, and ensemble integration hold in external cohorts that are similar to MA.27?

All models, the pre-trained *PREDICT v3*, the de novo ML (RSF, XGB) and the ensemble integration, presented with good discrimination and calibration. The discriminatory ability of *PREDICT v3* in our study was 0.738 (MA.27), 0.765 (SEER) and 0.701 (TEAM). This is worse than the published update by Grootes et al. where the AUC for 5-year survival in hormone positive cancer ranged between 0.831 and 0.861 [21]. Chen et al. [22] assessed *PREDICT v3* in a Chinese cohort of 5,424 women treated for nonmetastatic invasive breast cancer between 2010 and 2020 with an AUC for 5-year survival in the hormone positive sub-cohort was 0.789. Hsiao et al. [23] used SEER data from more than 860,000 patients (diagnosed 2000–2018) to validate *PREDICT v3* on US patients. In their study, the AUC for 5-year survival was 0.797 in the hormone positive sub-cohort. Calibration was more difficult to compare directly between these studies due to differences in calibration metrics and presentation. Visual comparison of calibration plots in the hormone positive sub-cohort, however, suggest that *PREDICT v3* was generally well-calibrated across these three studies [21–23]. If at all, there was a tendency to underestimate survival at the lower end of the survival prediction spectrum. In our study, this tendency was confirmed in MA.27 (ICI 0.042) and SEER (ICI 0.039) and appeared more pronounced than in literature. TEAM, in contrast, showed an opposite pattern with a clear tendency to overestimate survival (ICI 0.034). While not directly comparable, a very recent validation of *PREDICT v2.1* (not *PREDICT v3*) on Canadian patients with breast cancer diagnosed between 2004 and 2020 from Alberta achieved more similar *PREDICT v3* result to our study with an AUC of 0.78 and an ICI of 0.03 (mixed cohort with hormone positive and hormone negative cancer) [37].

The lower performance of *PREDICT v3* in our study and the one in Alberta reflect the challenge discussed in the introduction: MA.27 differs from the data used in the other studies, or more broadly, cohorts where pre-trained models are deployed may relevantly diverge from the original training data.

Transfer learning (i.e., *f-PREDICT v3*), de-novo RSF (but not XGB) and ensemble integration outperformed the stand alone pre-trained *PREDICT v3* in MA.27 and were similar in their performance. As model training was optimized for calibration, this was most pronounced in the reduction of ICI (*f-PREDICT v3* 0.005, RSF 0.003, ensemble 0.007 versus *PREDICT v3* 0.042, XGB 0.040). The discriminatory ability also improved with an AUC up to 0.799 (*f-PREDICT v3*) even though these differences would rather be considered negligible (less than 0.1 difference in AUC). The external validation of the MA.27-trained models yielded mixed results: while model validation on the SEER cohort confirmed these findings, the benefit of transfer learning, de-novo RSF and ensemble integration did not apply to TEAM where the pre-trained *PREDICT v3* outperformed all alternative approaches with, however, rather low calibration (ICI 0.039).

To come back to the questions posed: First, fine-tuning the pre-trained prognostic tool *PREDICT v3* to the MA.27 dataset led to a substantial improvement in performance, particularly in terms of calibration but also in terms of discrimination, compared to the pre-trained model alone. Second, state-of-the-art ML models trained directly on the MA.27 dataset have mixed results. RSF matches *f-PREDICT v3* and outperforms the stand-alone pre-trained model in terms of calibration but not in terms of discrimination; XGB showed the opposite pattern, with better discrimination but calibration closer to the stand-alone pre-trained model. Importantly, ML models, unlike (fine-tuned) *PREDICT-v3*, come with the advantage of handling missing values which can affect a non-trivial portion of the overall dataset (23.8-25.8% in MA.27). While the performance of RSF was lower for these individuals, it still achieved discrimination comparable to the stand-alone pre-trained *PREDICT v3* on individuals it could predict and demonstrated superior calibration. Third, the ensemble integration did not add further benefit over either approach alone but, again, came with the advantage of providing predictions for all patients which is not the case with *f-PREDICT v3*. And fourth, the observed benefits from fine-tuning and de-novo ML did extend to a similar SEER cohort. In this case, the ensemble appeared to be the best approach given its ability to handle missingness, its superior performance in both calibration and discrimination compared to *PREDICT-v3* and its comparable performance to *f-PREDICT v3*. In contrast, none of these benefits generalized to the TEAM cohort which may be due to involvement of multiple countries and variations in standards of care.

### 4.2. Implications for Practice

This study demonstrates that parameter-based transfer learning, de-novo ML training and ensemble integration can help to improve prognostication in situations where relevant information for *PREDICT v3* is lacking or a dataset shift is likely. Interestingly, these benefits can even generalize beyond the training cohort.

While missing information may not play a role in clinical prognostication where patients provide real-time information, it is commonly encountered when doing retrospective survival analyses. In such situations, de-novo ML training and ensemble integration can be good approaches for more reliable survival prediction. Beyond individual patient care, such models can be leveraged for emerging in-silico trial designs by simulating counterfactual outcomes and enabling virtual comparisons of treatment effects [38–40].

### 4.3. Limitations

This study has certain limitations which are mainly driven by the dataset's characteristics. There were some variables necessary for *PREDICT v3* entirely missing in the dataset (e.g. chemotherapy regimen in case of chemotherapy), so that assumptions were made where reasonably possible (see details in **Appendix A**). *PREDICT v3* may perform better in situations where such assumptions are not necessary.

The follow-up time in MA.27 was limited to 5 years. While 5-year survival represents an early and clinically relevant milestone to guide decision-making, screening and treatment developments have significantly improved outcomes in recent years, so that longer-term survival such as 10- or 15-year survival are becoming more relevant.

The methodological work on model improvement used the MA.27 clinical trial, which represents an older cohort. Nevertheless, the learnings would be applicable to more recent datasets and the same methods can be applied to more recent cohorts to improve contemporary prognostic model performance.

It is also relevant to note that prognostic application is different from predictive one [3]. The good performance of our model in terms of survival prognostication does not imply that the models can estimate individual-level treatment benefits. Causal inference methods must be leveraged to better interpret such functionality.

## 5. Ethics

This project has been approved by the Ottawa Health Science Network Research Ethics Board (protocol ID 20210803-01H) and by the Children's Hospital of Eastern Ontario Research Ethics Board (protocol 25/107X).

## 6. Acknowledgements

The authors would like to thank Paul Pharoah (Department of Computational Biomedicine, Cedars-Sinai Medical Center, Los Angeles, USA) for the kind provision of the *PREDICT v3* algorithm and the helpful comments on its use. We also would like to thank Daniel W. Rea (Cancer Research UK Clinical Trials Unit, University of Birmingham, United Kingdom) for his contribution to the TEAM study and all patients and staff involved in the MA.27 and TEAM study.

## 7. Author Contributions

Conceptualization, design, and analysis: LP, GRP, KY, LJ, FKD, MC, KEE; data collection and acquisition: ABB, LV, JH, MS, AL, LS, BEC, JMSB, KJT, JB, SLB, MS, CJHV, EMKK, LD, EM, AH, CM, MC; drafting manuscript: LP, KEE; review and editing: all authors.

## 8. Funding Statement

LP is funded by the Deutsche Forschungsgemeinschaft (DFG, German Research Foundation) – 530282197. KEE is funded by the Canada Research Chairs program through the Canadian Institutes of Health Research, and a Discovery Grant RGPIN-2022-04811 from the Natural Sciences and Engineering Research Council of Canada. The Ontario Institute for Cancer Research (OICR) is supported by funds from the Government of Ontario.

## 9. Competing Interests Statement

KEE was the scholar-in-residence at the Office of the Information and Privacy Commissioner of Ontario at the time of conducting this study.

## 10.  Data Availability

We used confidential healthcare data (MA.27 and TEAM) as well as accessible data from SEER for this study. Access to SEER can be requested under www.seer.cancer.gov.

# Transfer Learning and Machine Learning for Training Five Year Survival Prognostic Models in Early Breast Cancer: Appendix A

# 1.   Supplemental Background

## 1.1   Traditional Survival Analysis Methods

Modeling survival outcomes has a long history in statistics. The well-known Kaplan-Meier estimator [1] is one of the early foundational contributions to survival analysis and nonparametric statistics. It estimates the survival function directly from censored data, without any parametric assumption. The Kaplan-Meier estimator provides stepwise survival probabilities, allowing researchers to visualize survival functions and compare them across groups using statistical tests such as the log-rank test. However, it does not account for covariates, limiting its utility in multivariate analyses.

The Cox Proportional Hazards model [2], introduced under the proportional hazards assumption, represents a significant advancement as a semiparametric method to model survival outcomes. It estimates the hazard ratio associated with covariates while leaving the baseline hazard function unspecified. This flexibility allows the Cox model to handle complex survival data without distributional assumptions. However, the model relies on the proportional hazards assumption, which posits that the hazard ratios are constant over time. Violations of this assumption may limit its applicability in some datasets.

Parametric survival models have also been widely applied, leveraging specific assumptions about the distribution of survival times. Examples include the exponential, Weibull, log-normal, and log-logistic models [3]. These models provide precise estimates and enable extrapolation beyond the observed time range when the chosen distribution aligns with the underlying survival process. However, the parametric nature of these models makes them sensitive to misspecification of the distributional assumption.

Extensions of survival models include frailty models [4], which incorporate random effects to account for unobserved heterogeneity. Frailty models introduce a latent random variable, or *frailty,* that captures variability in survival outcomes not explained by observed covariates. Shared frailty models, for example, assume that individuals within the same group share a common frailty, leading to correlated survival times. These models are particularly useful for analyzing clustered data, where accounting for within-cluster dependence is crucial.

The Accelerated Failure Time (AFT) model [5] adopts a parametric approach to survival analysis by modeling survival time directly as a multiplicative function of covariates. In contrast to the Cox model, which focuses on hazard ratios, the AFT model examines how covariates accelerate or decelerate survival time. AFT models rely on assumptions about the distribution of survival times, commonly employing Weibull, log-normal, or log-logistic distributions. This approach provides an intuitive interpretation stemming from the distributional assumption, especially in many biological contexts where the timing of events is of primary interest. There is evidence that AFT models may naturally align with certain biological processes [6], particularly in scenarios where the proportional hazards assumption is violated.

## 1.2   Machine Learning in Survival Analysis

In recent years, the rapid advancement of machine learning (ML) research has led to the development of numerous approaches to predict survival outcomes. These methods leverage various algorithms to model survival data, accommodating censored observations and complex relationships among covariates. Among the prominent ML techniques adapted for survival analysis are tree-based methods, neural networks, support vector machines, and Bayesian approaches.

Tree-Based Methods, such as Random Survival Forests [7] and gradientboosting trees like XGBoost [8], have been effectively extended to handle survival data. Random Survival Forests are a nonparametric tree-based ensemble learning method. At each node of a tree, the best split is determined based on survival-specific criteria, for example, the log-rank statistic. This criterion aims to maximize the difference in survival probabilities between child nodes, thereby identifying the most informative splits for predicting survival outcomes. Similarly, gradient-boosting frameworks like XGBoost have been adapted for survival analysis by incorporating survival-specific loss functions. XGBoost extends gradient-boosting trees using objectives based on the Cox proportional hazards model or accelerated failure time models.

Neural Networks, recognized as universal approximators [9], [10], have been extensively applied in survival modeling to capture complex, nonlinear relationships and interactions among covariates. Various deep learning (DL) architectures have been adapted for survival analysis. For example, DeepSurv [11] extends the Cox proportional hazards model by employing a multilayer perceptron to model the hazard function. The network is trained by maximizing the Cox partial likelihood, allowing it to learn the underlying hazard relationships from data. DeepHit [12] combines DL with competing risks by modeling the joint distribution of survival times and event types. It utilizes a discrete-time hazard function and optimizes a loss function that accounts for both the event type and the time of occurrence. In addition, numerous approaches have been developed that use deep neural networks to model survival outcomes (see, for example, [13], [14], [15], [16], [17], [18]). Despite their flexibility and strong predictive performance, neural networks in survival analysis present several challenges. They often lack interpretability, making it difficult to understand the influence of individual covariates on survival outcomes. Furthermore, neural networks are prone to overparameterization, which can lead to overfitting, especially in scenarios with limited sample sizes. Effective neural network training typically requires large datasets to capture the underlying patterns. In addition, the performance of neural networks can be highly sensitive to hyperparameter settings, initializations, and optimization algorithms, leading to variability in performance.

Support Vector Machines (SVM) have been adapted for survival analysis to handle time-to-event data by employing either ranking or regression constraints [19], [20]. SVMs offer several advantages in survival analysis. They are particularly effective when applied to high-dimensional datasets and can capture complex, nonlinear relationships between covariates and survival outcomes through the use of kernel functions. However, SVMs have limitations. They lack straightforward interpretability of model parameters, similar to other kernel-based methods. The performance of SVMs is highly dependent on the choice of kernel and hyperparameters, necessitating careful tuning. Furthermore, SVMs can be sensitive to variations in input data, potentially leading to instability in predictions. Their effectiveness is also heavily impacted by censoring.

Bayesian Approaches to survival analysis offer a probabilistic framework that allows the incorporation of prior knowledge in the model parameters [21], [22]. Bayesian methods offer several advantages, including the ability to incorporate prior knowledge, which is particularly useful in scenarios with limited data and prior knowledge. The Bayesian framework naturally provides credible intervals for model parameters and survival predictions, facilitating more comprehensive inferences. Additionally, Bayesian models are highly flexible, accommodating various types of censoring and extending to handle competing risks and time-dependent covariates. Bayesian methods come with several challenges. Careful selection of prior distributions is crucial to avoid bias, and inappropriate priors can lead to misleading inferences, especially in a data-scarce scenario. In addition, specifying appropriate models that accurately reflect the underlying survival mechanisms can be challenging and may require substantial domain expertise.

ML approaches have been increasingly developed for breast cancer prognostication in the past few years [23], [24], [25]. In 5-year survival prediction, these tools are most often based on SEER data with a dataset size of more than 10,000 patients [24]. Applied ML approaches are typically decision trees, neural networks and SVM and can present with promising discrimination. For example, in [26], RSF, SVM and XGB based survival models were compared against the Cox Proportional Hazards model. In this study, XGB outperformed the other models in terms of discrimination. Another study [27] reported comparable results in terms of discrimination between RSF and the Cox Proportional Hazards model. As shown by a recent review, however, most studies with ML survival models lack a comprehensive performance evaluation including discrimination and calibration and are often not externally validated [24]. Also, various models treat survival as a binary classification problem thereby ignoring the time-to-event nature of that data [28], [29]. These shortcomings may be an explanation for their limited applicability in clinical practice.

### 1.3 Transfer Learning in Survival Analysis

Transfer learning, similar to ML, has gained increasing attention in recent years. The idea is that a new task (i.e., target domain) can be more easily learned by transferring knowledge from a related task (i.e., source domain) that has already been learned [30]. In the context of DL and neural networks, transfer learning is particularly interesting as large datasets are required but not always available or accessible for training. This is very valuable in domains, such as healthcare, where resources are typically scarce and large datasets difficult to access [31], [32], [33], [34], [35]. There are multiple different approaches to transfer learning but most of them can be categorized into four groups: instance-based, feature-based, parameter-based or relational transfer learning [30]. Instance-based transfer learning re-weights the source data to align better with the target distribution, feature-based transfer learning re-weights the source variables or identifies a common latent representation for source and target variables, parameter-based transfer learning re-uses the parameters from a model trained on source data and relational transfer is the transfer via modeled relationships between source and target.

In survival analysis, as with other tasks in medical research, transfer learning can help with challenges such as heavy censoring or small sample sizes. Instance-based transfer learning for optimized survival analysis has been proposed, for example, in [36] where individuals were re-weighted to better align with the target distribution. Another instance-based example has been applied in [37] where data from biologically similar cancer entities was selected to improve survival models. However, access to the source data for instance re-weighting may not always be feasible due to privacy concerns or intellectual property limitations. Parameter-based transfer learning is a practical alternative in such scenarios. Instead of re-weighting the instances of the source data itself, the parameters of the pre-trained model are fine-tuned to the target data. In lung cancer survival prediction, for example, Zhu et al. fine-tuned a pre-trained DL prediction model on their target cohort and demonstrated its superiority against the model without fine-tuning [38]. In another example, survival analysis based on cancer transcriptomics data performed best in a DL approach with parameter-based transfer learning compared to more traditional models [39]. Similarly, parameter-based transfer learning added value in survival analysis of pancreatic ductal adenocarcinoma based on medical imaging [40].

### 1.4 The Pre-Trained Model *PREDICT* in Breast Cancer Survival Analysis

*PREDICT* is a tool for survival prognostication in breast cancer that has been widely adopted in practice. It was originally fitted on 5,232 breast cancer cases from the UK East Anglia Cancer Registration and Information Centre diagnosed in 1999-2003 [12], updated recently to *PREDICT v3* with patients diagnosed between 2000 and 2017 [14] and validated on several cohorts around the world [15,16]. This version has been developed by Grootes, Wishart, and Pharoah [41] using a cohort from the UK. The total number of patients involved in development and validation exceeded 132,000. The model demonstrated robust discrimination (AUCs between 0.78 and 0.82) and excellent calibration, with predicted mortality within 10% of the observed values, thus correcting the over-prediction observed with the earlier version (*PREDICT v2.1*).

Chen et al. [42] assessed *PREDICT v3* in a Chinese cohort of 5,424 women treated for nonmetastatic invasive breast cancer between 2010 and 2020. Calibration was evaluated using quantile-based calibration graphs and chi-squared goodness-of-fit tests; The overall 5-year survival predicted by the model differed from the overall survival observed by approximately −2% and nearly 0% in patients with ER negative. However, the model underestimated 5-year survival by approximately 9% in patients older than 75 years and by 5.8% in those with micrometastases, while it overestimated survival in patients who later developed distant metastases. Discrimination was similar to that of v2.2, with an overall AUC around 0.756.

In [43], a US validation study using SEER data from more than 860,000 patients (diagnosed 2000–2018) examined *PREDICT v3*. Calibration was assessed by comparing observed versus predicted outcomes across risk quintiles, and the model's 10- and 15-year mortality estimates were within 5–8% of the observed values for both ER-positive and ER-negative patients. The AUC values were approximately 0.769 for ER-positive and 0.738 for ER-negative cases at 10 years. Notably, subgroup analysis revealed that the model over-predicted mortality in non-Hispanic Asian patients with ER-negative disease and under-predicted mortality in non-Hispanic Black patients with ER-positive disease, suggesting that recalibration may be needed for these groups.

The earlier version, *PREDICT v2.1,* was also recently validated in a cohort from Alberta, Canada, diagnosed between 2004 and 2020 [44]. Despite being the earlier version, it achieved good discrimination with AUC values of 0.78 for 5-year and 0.73 for 10-year survival. In comparison, an ML approach using Random Survival Forests demonstrated lower discrimination (0.67 and 0.64 respectively) and calibration.

## 2. Supplemental Methods

### 2.1 Outcome Re-Balancing

The Random Over-Sampling Examples (ROSE) technique can ensure a more balanced and representative distribution of outcome [1] and was tested during model training in this study. For ROSE, missingness was encoded as dummy variable (in numerical variables) or as separate category (in categorical variables) to preserve its structure. This was reversed in a post-processing step. We implemented ROSE via the R library ROSE [45]. Results are presented in the supplemental results section below.

### 2.2 Machine Learning Survival Models

In light of the advantages and disadvantages described in the supplemental background, we opted to use tree-based methods to model survival in the MA.27 dataset. This decision is informed by several key characteristics of the data and the inherent strengths of tree-based algorithms.

1. Predominance of Categorical Covariates: The MA.27 dataset primarily consists of categorical covariates. Tree-based methods, such as Random Survival Forests and Gradient Boosting Machines, are inherently adept at handling categorical variables without extensive preprocessing.

2. High Proportion of Censored Observations and Limited Number of Observations: With over 95% of the patients not experiencing the event within the 5-year observation period and approximately 7000 datapoints in total, the dataset presents significant challenges and makes deep learning approaches less viable due to their data-intensive nature and complex parameterization needs.

3. Limited Concern for Nonlinearity: Tree-based models effectively capture nonlinear relationships, provided they are not excessively complex. In the MA.27 dataset, the limited number of continuous variables minimizes concerns about nonlinearity. Therefore, methods such as support vector machines and neural networks, which are designed to handle more intricate nonlinear interactions, offer limited advantages in this context.

4. Natural Handling of Missing Data: Tree-based algorithms manage different missingness types as mentioned in the section on Data Management. This makes them also ideal for developing web-based prediction tools since they can seamlessly manage incomplete user provided data while ensuring robust predictions.

In summary, the decision to leverage tree-based methods for survival prognostication in the MA.27 dataset was underpinned by the dataset's characteristics. Random Survival Forests (RSF), an ensemble method tailored for survival analysis, and Extreme Gradient Boosting (XGB), an efficient and scalable implementation of gradient boosting machines with a survival-specific loss function, were leveraged in this study. RSF was implemented via the R package randomForestSRC [47]; XGB via the R package xgboost [48].

To enhance the representation of the continuous variables (age and maximum tumor size) for these ML models, we applied normalization and constructed Bernstein polynomials of degree 3.

### 2.2.1 Hyperparameter Tuning

Hyperparameter tuning for these models was executed using a zero-order search methodology (grid search) to optimize model performance across all components using the MA.27 testing subset (i.e., data B). The following hyperparameters were considered:

|     | Hyperparameter | Explanation | Values considered |
| --- | --- | --- | --- |
| RSF | ntree | Number of trees in the RSF | 500; 1000; 1500 |
| RSF | mtry | Number of variables sampled at each split | 3; 4; 6 |
| RSF | nodesize | Minimum terminal node size | 3; 5; 10; 15 |
| RSF | splitrule | Splitting criterion | logrank; logrankscore |
| XGB | eta | Learning rate | 0.05; 0.1 |
| XGB | max_depth | Maximum depths of trees | 2; 5 |
| XGB | subsample | Subsample ratio of training instances | 0.6; 1 |
| XGB | colsample_bytree | Subsample ratio of columns when constructing each tree | 0.6; 1 |
| XGB | lambda_vals | L2 regularization coefficient on leaf weights | 0.05; 0.1 |
| XGB | nrounds | Number of boosting rounds | 500 |

**Table 1. Hyperparameters Considered in Grid Search When Tuning ML Survival Models.**

## 2.3 Pre-Trained Model and Transfer Learning

*PREDICT v3* is publicly available under [49]. We adjusted the code to flag potential contradictory inputs (e.g. HER2 therapy but HER2 negativity), automatically impute detection mode by age and screening practices and infer mean heart dose by tumor laterality. The code was further vectorized to allow for more efficient calculation.

To fine-tune *PREDICT v3* for the MA.27 cohort, we optimized its 26 parameters using the MA.27 training subset (i.e., parameter-based transfer learning). This optimized version is referred to as *f-PREDICT v3.* The parameter search was conducted via the gradient-free optimization algorithm Nelder-Mead [50] which is more robust in settings where differentiability cannot easily be assumed. This is a local optimization approach and was implemented via the R package STAT [51].

### 2.4 Variable Mapping of MA.27, SEER and TEAM

*PREDICT v3* requires certain information about the patient to predict their survival. While some missing variables can be handled internally by the model, the absence of certain key predictors prevents the tool from giving a valid survival estimate. It should further not be used in women with DCIS/LCIS only or in women with metastatic disease. The MA.27 and TEAM eligibility criteria were in line with these requirements; SEER data was selected to match these eligibility criteria.

MA.27 collected various information including demographic, clinicopathological and treatment-related variables but did not provide all of the necessary information at the desired level of granularity, or in some cases, at all. In **Table 2**, we give details about the *PREDICT v3* input variables and their availability in MA.27 and, if applicable, assumptions we could take based on background knowledge about the study population. **Table 3** and **Table 4** give these details for SEER and TEAM respectively.

| Variable in *PREDICT v3* | Mapping to MA.27 | Missingness |
|---|---|---|
| Year of Diagnosis | Year of diagnosis was unknown for participants in the MA.27 study. We assumed diagnosis to be close to enrollment since the treatment timeline for breast cancer patients is well established in clinical practice and the initiation of hormone therapy typically occurs within a few months following diagnosis and surgery. We therefore used the start of recruitment (i.e., 2003) as year of diagnosis. We did not have the precise year of enrollment and year has a positive effect on survival, so that we may have underestimated survival (see limitations in main manuscript). | 100.0% |
| Age in Years | Age in years was available for all patients in MA.27. | 0.0% |
| Postmenopausal Status | All patients in MA.27 were postmenopausal (see inclusion criteria [52]). | 0.0% |
| Smoking Status | As smoking status is required by *PREDICT v3* but was not available in MA.27, we assumed smoker status as 0 (i.e., non-smoker). However, at the time of MA.27 (i.e., 2003), 21% of Canadians were current smokers [53] and smoking can influence treatment outcome (see limitations in main manuscript). We also trained a model based on the assumption of a positive smoker status for all patients without any differences in performance measurements so that we decided to assume non-smokers. | 100.0% |
| Estrogen Receptor Status | Information on estrogen receptor status was available for all patients in MA.27. | 0.0% |
| Progesterone Receptor Status | Information on progesterone receptor status was not available for all patients in MA.27 and could not be inferred for those patients | 2.0% |

| Variable in *PREDICT v3* | Mapping to MA.27 | Missingness |
|---|---|---|
| | using background information. This information is not necessarily required by *PREDICT v3* but can be left as missing. | |
| HER2 Status | The MA.27 dataset did not provide details on the HER2 status but includes a variable, indicating Herceptin (trastuzumab) usage. This variable was introduced in 2005, following the publication of positive results showing efficacy of trastuzumab in HER2-positive early breast cancer patients. While we can assume that patients with trastuzumab usage are HER2 positive, the situation is more complex for patients who did not receive trastuzumab: from 2005 onwards, it is very likely that these patients are HER2 negative, as trastuzumab would have been part of their standard treatment, if they were positive. Before 2005, however, this assumption does not hold as trastuzumab was not widely used for HER2 positive patients. Since we did not have the year of diagnosis or enrollment, we could not make a distinction. While *PREDICT v3* accepts missing information as an input, we decided to assume HER2 negativity for patients where information about trastuzumab was not indicated (see limitations in main manuscript). This may better reflect clinical reality since around 85% of breast cancers are HER2 negative [54]. | 100.0% |
| Ki-67 Status | Information on Ki-67 was not available in MA.27 but missingness could be internally handled by *PREDICT v3*. | 100.0% |
| Tumor Size in mm | Information on tumor size in mm was available for most patients. This information could not be inferred using background information but is required by *PREDICT v3* so that survival could not be estimated for these patients using *PREDICT v3*. | 20.7% |
| Tumor Grade | This reflects the pathological tumor grading. MA.27 used a grading from 1 (well-differentiated) to 3 (poorly differentiated). This information could not be inferred using background information but is required by *PREDICT v3* so that survival could not be estimated for these patients using *PREDICT v3*. | 21.9% |
| Mode of Detection | This variable was not available in the MA.27 dataset. In line with recent *PREDICT v3* validation studies [43], however, we added the following imputation process based on age and screening practices to *PREDICT v3*. For those within the standard screening programme age range (between 50-75 years), we set the mode of detection value to 0.5 to account for the proportion of eligible women who have regular screening in this population and the proportion of cancers in women who are interval cancers in those who have regular screening. For those outside the screening programme age range, mode of detection was set as symptomatic. | 100.0% |
| Number of Nodes | *PREDICT v3* requires the exact number of nodes involved. MA.27 provided lymph node information through the TNM classification (see limitations in the main manuscript) so that we used the following approximations: N0 as 0 nodes, N1 as 2 nodes, N2 as 7 nodes and N3 | 0.1% |

| Variable in *PREDICT v3* | Mapping to MA.27 | Missingness |
|---|---|---|
| | as 10 nodes. Survival could not be estimated for patients with missing TNM classification using *PREDICT v3*. | |
| Micrometastases | Information on micrometastases is relevant in case of one positive node which would set the value of node to 0.5 instead of 1. This was not relevant since the approximation for the number of nodes would not one positive node. | 100.0% |
| Radiotherapy | The MA.27 dataset did provide details on radiotherapy. Survival could not be estimated for patients with missing information on radiotherapy using *PREDICT v3*. | 28.9% |
| Mean Heart Dose for Radiotherapy | In case of radiotherapy, the mean heart dose is required for *PREDICT v3*. This was not available in MA.27 but *PREDICT v3* provides the following instruction: "If you are unsure, use 0 for cancer on the right-hand side and 2 for cancer on the left-hand side." MA.27 provided tumor laterality for all patients who underwent radiotherapy so that we followed these instructions. | 100.0% |
| Hormone Therapy | All patients in MA.27 received hormone therapy (see clinical trial details [52]). | 0.0% |
| Chemotherapy | While MA.27 provided complete information whether or not chemotherapy was part of the treatment, details regarding whether it was standard anthracycline or taxanes / high-dose anthracycline are missing. Taxanes (paclitaxel and docetaxel) have been introduced as chemotherapy for metastatic breast cancer in 1994 and 1996 by the FDA [55]. However, there was still an on-going debate on their use in the adjuvant setting [56], [57] at the time of MA.27. A Canadian Medical Guideline on adjuvant systemic therapy for women with node-negative breast cancer around that time (2001 update) concluded that taxanes (more precisely AC-Taxol) had not been evaluated in node negative disease at that time. Similarly, the Canadian guideline on adjuvant systemic therapy for women with node-positive breast cancer (2001 update) mentioned that this is a field of research and that participation in trials investigating taxanes should be encouraged. MA21 was initiated by the National Cancer Institute of Canada-ClinicalTrials Group in 2000 based on evidence pointing towards a benefit of taxanes [58]. The rationale for the study [58] highlights that based on results of the CALGB 9344 study, AT/T can be considered as standard therapy for node-positive patients. Consequently, in MA21 the two standard regimens they compared against were CEF (Cyclophosphamid, Epirubicin and 5-Fluorouracil) and AC/T (doxorubicin and cyclophosphamide, followed by paclitaxel) whereby the first one can be considered as standard-dose anthracycline-based therapy and the second one as taxane-based therapy. The regimen they were interested in was EC/T (epirubicin and cyclophosphamide, followed by paclitaxel) which is a high-dose anthracycline and taxane-based therapy, and | 0.0% |

| Variable in *PREDICT v3* | Mapping to MA.27 | Missingness |
|---|---|---|
| | demonstrated superiority over AC/T but no superiority or inferiority over CEF [59]. Based on the context surrounding MA21 and insights from our co-authors (MC) who has been practicing in this field for decades, it can be concluded that taxanes/high-dose anthracycline-based therapy can be assumed for patients at the time of MA.27. | |
| Trastuzumab Therapy | The MA.27 dataset introduced trastuzumab usage as variable in 2005, following the publication of positive results showing efficacy of trastuzumab in HER2-positive early breast cancer patients. As a result, the majority of patients had no information on trastuzumab usage, most of whom were likely enrolled prior to 2005, when trastuzumab was not yet standard of care. Given this historical context, we assumed no trastuzumab use in cases with missing data. | 74.7% |
| Bisphosphonate Use | The original publication mentions concurrent bisphosphonate use in 809/7,576 (11%), but this variable was absent in the MA.27 data we possess. Since bisphosphonate in non-metastatic breast cancer is rather unusual and not standard treatment, we assumed no bisphosphonate therapy. | 100.0% |

**Table 2. Mapping MA.27 Variables to *PREDICT v3*.** Missingness is indicated in the original variables before imputing as described in the table. The total number of individuals from MA.27 was 7,563.

| Variable in *PREDICT v3* | Mapping to SEER | Missingness |
|---|---|---|
| Year of Diagnosis | Year of diagnosis was known in SEER and those patients that had the same year of diagnosis as MA.27 were selected (i.e., 2003) | 0.0% |
| Age in Years | Age in years was available for all patients. | 0.0% |
| Postmenopausal Status | Postmenopausal status was not available in SEER but could be approximated by the age of the patients. An age below 45 years was considered as premenopausal. | 100.0% |
| Smoking Status | As smoking status is required by *PREDICT v3* but was not available in SEER, we assumed smoker status as 0 (i.e., non-smoker). The same limitation applies as for the MA.27 data in this regard. | 100.0% |
| Estrogen Receptor Status | Information on estrogen receptor status was available for all patients in our SEER subset. | 0.0% |
| Progesterone Receptor Status | Information on progesterone receptor status was not available for all patients. This information is not necessarily required by *PREDICT v3* but can be left as missing. | 3.8% |
| HER2 Status | SEER did not provide details on the HER2 status. We assumed HER2 negativity in line with the assumptions in the context of MA.27. | 100.0% |
| Ki-67 Status | Information on Ki-67 was not available in SEER but missingness could be internally handled by *PREDICT v3*. | 100.0% |

| Variable in *PREDICT v3* | Mapping to SEER | Missingness |
|---|---|---|
| Tumor Size in mm | Information on tumor size in mm was available for most patients. Survival could not be estimated for patients with missing information on tumor size using *PREDICT v3*. | 6.1% |
| Tumor Grade | This reflects the pathological tumor grading. SEER used a grading from 1 (well-differentiated) to 4 (undifferentiated, anaplastic). We mapped grade 4 to grade 3, reflecting their shared classification as poorly differentiated tumors. Survival could not be estimated for patients with missing information on tumor grade using *PREDICT v3*. | 7.2% |
| Mode of Detection | This variable was not available in the SEER data. We inferred mode of detection as done for MA.27. | 100.0% |
| Number of Nodes | *PREDICT v3* requires the exact number of nodes involved. In contrast to MA.27, this information is available for most patients in SEER. Survival could not be estimated for patients with missing information on nodes using *PREDICT v3*. | 10.2% |
| Micrometastases | Information on micrometastases is relevant in case of one positive node which would set the value of node to 0.5 instead of 1. We assumed that no micrometastases were present in the case of 1 positive node. This assumption affected 12.2% of all patients. | 100.0% |
| Radiotherapy | The SEER dataset did provide details on radiotherapy. We aggregated any form of radiotherapy such as beam radiation, radioactive implants, radioisotopes or no specified method under radiotherapy. | 0.0% |
| Mean Heart Dose for Radiotherapy | In case of radiotherapy, the mean heart dose is required for *PREDICT v3*. We inferred mode of detection as done for MA.27, but tumor laterality was missing in 0.01% where we assumed 1 Gray. | 100.0% |
| Hormone Therapy | Hormone therapy was not directly available in SEER, but it could be assumed for all hormone receptor positive patients. | 100.0% |
| Chemotherapy | While SEER similar to MA.27 provided information whether or not chemotherapy was part of the treatment, details regarding whether it was standard anthracycline or taxanes / high-dose anthracycline are missing. We made the same assumptions as for MA.27. | 100.0% |
| Trastuzumab Therapy | SEER did not provide details on trastuzumab therapy nor on HER2. Since we assumed HER2 negativity, we also assumed no trastuzumab therapy. | 100.0% |
| Bisphosphonate Use | SEER did not provide information on bisphosphonate use but based on the same considerations as MA.27, we assumed no bisphosphonate therapy. | 100.0% |

**Table 3. Mapping SEER Variables to *PREDICT v3*.** The total number of individuals from SEER was 27,064.

| Variable in *PREDICT v3* | Mapping to TEAM | Missingness |
|---|---|---|
| Year of Diagnosis | Patient recruitment in TEAM covered a period from 2000 to 2006. Again, year of diagnosis can be assumed to be close to enrollment so that, similar to MA.27, the year of diagnosis was set to 2003 in all cases. | 100% |
| Age in Years | Age in years was available for all patients. | 0.00% |
| Postmenopausal Status | All patients were postmenopausal in TEAM (see eligibility criteria in [60]) | 0.00% |
| Smoking Status | As smoking status is required by *PREDICT v3* but was not available in TEAM, we assumed smoker status as 0 (i.e., non-smoker). The same limitation applies as for the MA.27 data in this regard. | 100% |
| Estrogen Receptor Status | Information on estrogen receptor status was available for all but one patient in TEAM. Survival could not be estimated for this patient. | 0.03% |
| Progesterone Receptor Status | Information on progesterone receptor status was not available for all patients. This information is not necessarily required by *PREDICT v3* but can be left as missing. | 12.9% |
| HER2 Status | Similar to MA.27, the HER2 status was not available for all participants of the TEAM trial. In line with MA.27, we decided to assume HER2 negativity for patients where information about HER2 was not indicated (see limitations in main manuscript). This may better reflect clinical reality since around 85% of breast cancers are HER2 negative [54]. | 6.6% |
| Ki-67 Status | Information on Ki-67 was available in TEAM as percentage and was mapped to positive in case of > 10%. Missingness could be internally handled by *PREDICT v3*. | 3.2% |
| Tumor Size in mm | Information on tumor size in mm was available for most patients. Survival could not be estimated for patients with missing information on tumor size using *PREDICT v3*. | 1.6% |
| Tumor Grade | This reflects the pathological tumor grading. Survival could not be estimated for patients with missing information on tumor grade using *PREDICT v3*. | 4.8% |
| Mode of Detection | This variable was not available in the TEAM data. We inferred mode of detection as done for MA.27. | 100% |
| Number of Nodes | *PREDICT v3* requires the exact number of nodes involved. In contrast to MA.27, this information is available for most patients in TEAM. Survival could not be estimated for patients with missing information on nodes using *PREDICT v3*. | 16.1% |
| Micrometastases | Information on micrometastases is relevant in case of one positive node which would set the value of node to 0.5 instead of 1. We assumed that no micrometastases were present in the case of 1 positive node. This assumption affected 25.6% of all patients. | 100% |
| Radiotherapy | The TEAM data did provide information on whether radiotherapy was applied. Survival could not be estimated for patients with missing information on radiotherapy using *PREDICT v3*. | 0.2% |

| Variable in *PREDICT v3* | Mapping to TEAM | Missingness |
|---|---|---|
| Mean Heart Dose for Radiotherapy | In case of radiotherapy, the mean heart dose is required for *PREDICT v3*. This was not available in TEAM nor information on tumor laterality to follow the instructions of *PREDICT v3* as mentioned above. Following the SEER procedure, we set the dose to 1 Gray in the case of radiotherapy which affected 64.9% of the patients. | 100% |
| Hormone Therapy | All patients in TEAM received hormone therapy (see clinical trial details [60]). | 0.0% |
| Chemotherapy | TEAM provided information on whether or not chemotherapy was part of the treatment, but did not provide consistent information about the regimen. To align with the trained models on MA.27, we set all chemotherapy applications to standard anthracyclines. | 0.1% |
| Trastuzumab Therapy | TEAM did not provide details on trastuzumab therapy. We assumed no trastuzumab therapy in case of HER2 negativity, and trastuzumab therapy in case of positivity. | 100% |
| Bisphosphonate Use | TEAM did not provide information on bisphosphonate use but based on the same considerations as MA.27, we assumed no bisphosphonate therapy. | 100% |

Table 4. Mapping TEAM Variables to *PREDICT v3*. The total number of individuals from TEAM was 3,825.

## 2.5 Ensemble Integration

The ensemble combined predictions from three individual models: *f-PREDICT v3*, RSF and XGB. The convex combination coefficients for ensemble weighting were treated as continuous hyperparameters and the best combination was searched via Bayesian optimization. Bayesian optimization was implemented via the R package rBayesianOptimization [61]. Since *f-PREDICT v3* cannot generate survival estimates when certain inputs are missing, the ensemble was designed to fall back to the remaining models, RSF and XGB, in such cases. As a result, the ensemble could still provide robust predictions even when there was incomplete input data.

## 3. Supplemental Results

### 3.1 External Evaluation Plots

External evaluation was conducted on data from the US SEER program and on the clinical trial dataset TEAM. ICI and AUC values are presented in the main manuscript. **Figure 1** and **Figure 2** provide the graphical illustration of calibration and discrimination for SEER; **Figure 3** and **Figure 4** for TEAM. The illustrations confirm the mixed findings from the main manuscript: while *PREDICT v3* performed similar across all datasets, MA.27 tailored transfer learning, de-novo ML and ensemble integration did outperform the pre-trained models in terms of discrimination and calibration in SEER but not in TEAM. In TEAM, all models consistently underestimated survival.

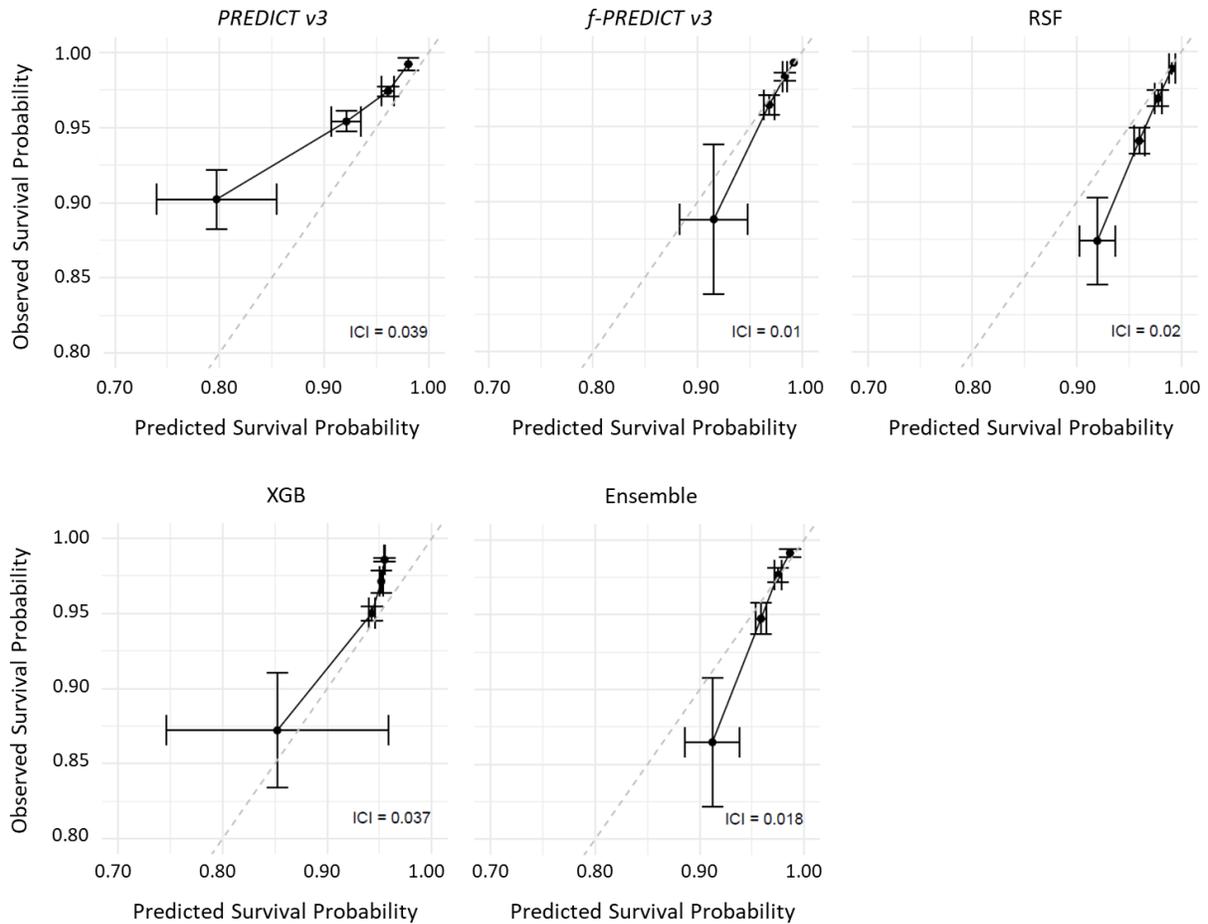

**Figure 1. Calibration Plots for External Evaluation on SEER.** Calibration plots are illustrated for the baseline model *PREDICT v3* as well as the four models enhanced by transfer learning (i.e., *f-PREDICT v3*), ML (RSF and XGB) and ensemble integration. The diagonal dashed line indicates perfect calibration. Observed probabilities were smoothed using a hazard regression-based method [62]. Observations were then divided into four quartiles based on their predicted probabilities. Both predicted and observed probabilities were trimmed to exclude extreme values beyond the 10[th] to 90[th] percentile, and mean predicted and observed survival probabilities were calculated in each quartile. The horizontal and vertical error bars reflect the standard deviations of these quartile-wise means. The calculated ICI is given for each model.

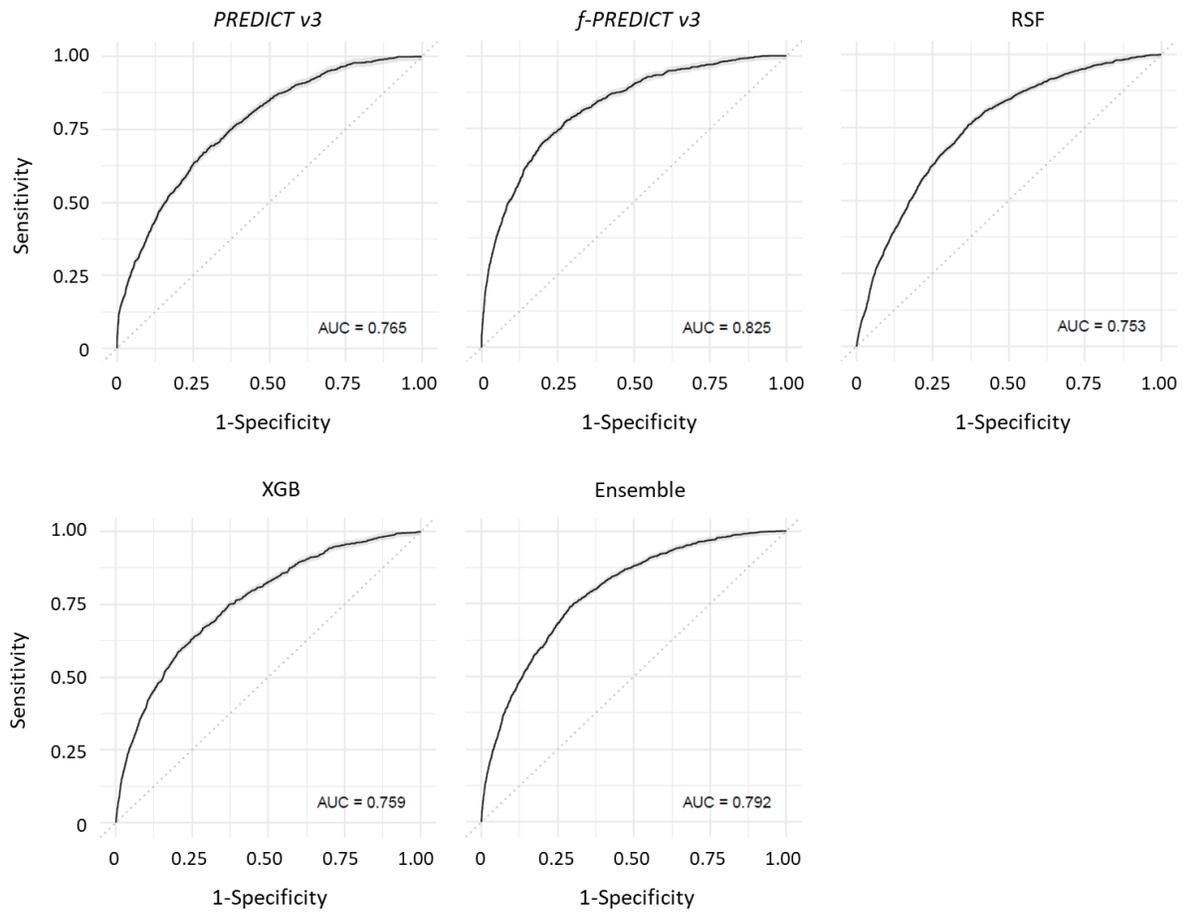

**Figure 2. Receiver Operating Characteristics (ROC) Curves for SEER**. ROC curves are illustrated for the baseline model *PREDICT v3* as well as the four models enhanced by transfer learning (i.e., *f-PREDICT v3*), ML (RSF and XGB) and ensemble integration. The diagonal dashed grey line indicates discrimination of a random guess. The calculated AUC value is given for each model.

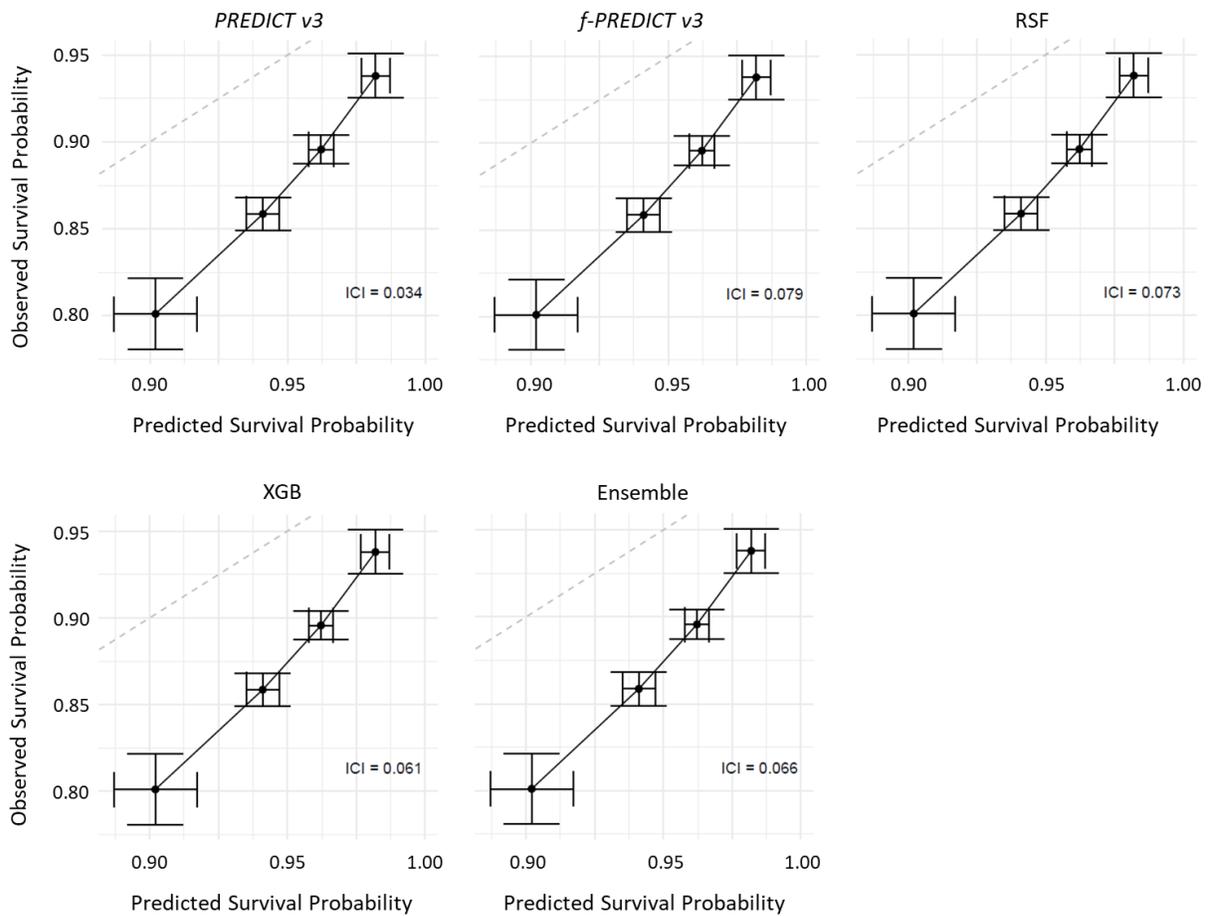

**Figure 3. Calibration Plots for External Evaluation on TEAM.** Calibration plots are illustrated for the baseline model *PREDICT v3* as well as the four models enhanced by transfer learning (i.e., *f-PREDICT v3*), ML (RSF and XGB) and ensemble integration. The diagonal dashed line indicates perfect calibration. Observed probabilities were smoothed using a hazard regression-based method [62]. Observations were then divided into four quartiles based on their predicted probabilities. Both predicted and observed probabilities were trimmed to exclude extreme values beyond the 10$^{th}$ to 90$^{th}$ percentile, and mean predicted and observed survival probabilities were calculated in each quartile. The horizontal and vertical error bars reflect the standard deviations of these quartile-wise means. The calculated ICI is given for each model.

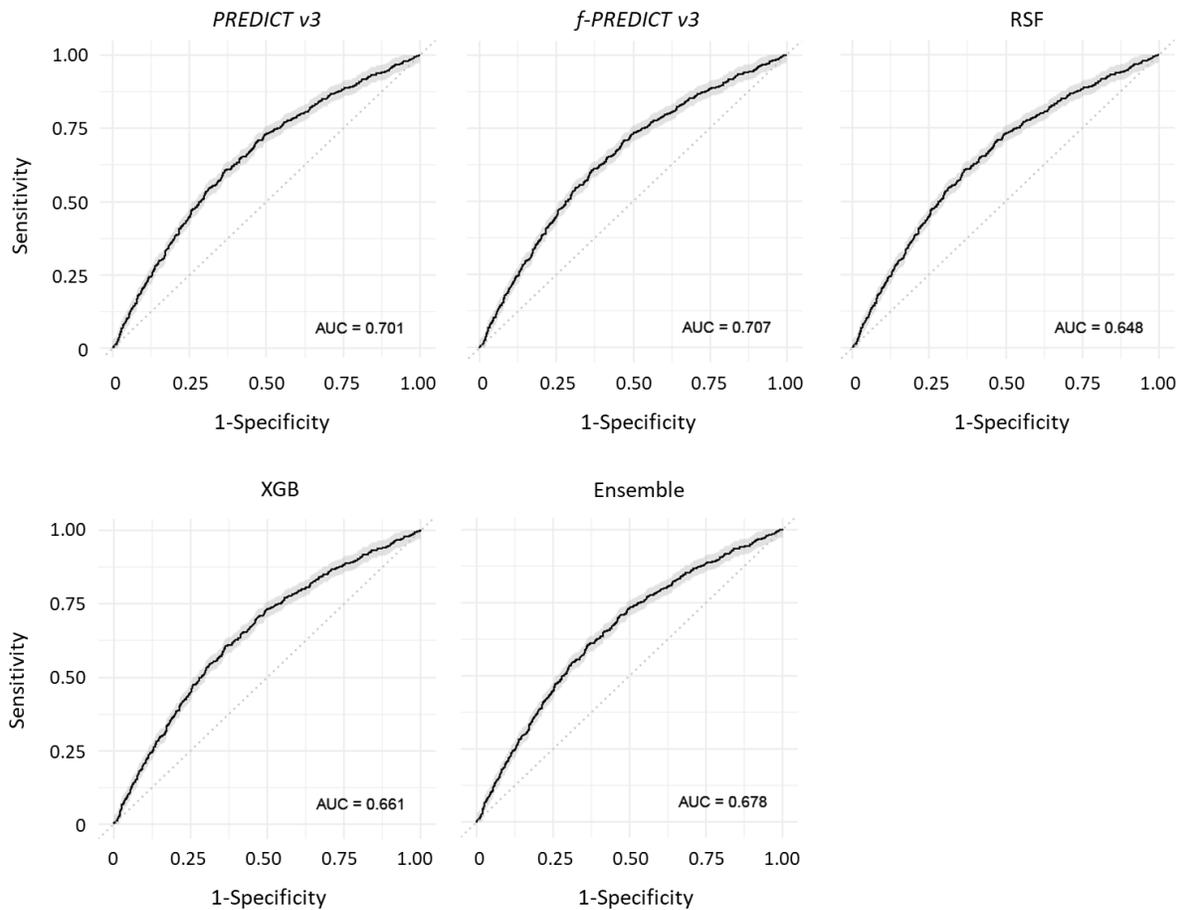

**Figure 4. Receiver Operating Characteristics (ROC) Curves for TEAM**. ROC curves are illustrated for the baseline model *PREDICT v3* as well as the four models enhanced by transfer learning (i.e., *f-PREDICT v3*), ML (RSF and XGB) and ensemble integration. The diagonal dashed grey line indicates discrimination of a random guess. The calculated AUC value is given for each model.

### 3.2    Re-Balancing For Model Training

As in the main manuscript, the models were optimized for ICI and re-balancing was tested. Results of training with re-balancing in **Table 5** show that calibration deteriorated when leveraging ROSE during training.

| Model | Calibration (ICI) | | Discrimination (AUC) | |
|---|---|---|---|---|
| | Median | IQR | Median | IQR |
| *PREDICT v3* | 0.042 | 0.039-0.049 | 0.738 | 0.719-0.770 |
| *f-PREDICT v3* | 0.334 | 0.328-0.335 | 0.812 | 0.778-0.838 |
| RSF | 0.247 | 0.232-0.266 | 0.784 | 0.764-0.798 |
| XGB | 0.554 | 0.549-0.565 | 0.792 | 0.786-0.802 |
| Ensemble | 0.287 | 0.238-0.318 | 0.787 | 0.768-0.807 |

**Table 5. Discrimination (AUC) and Calibration (ICI) When Testing Re-Balancing.** AUC and ICI values were calculated on the validation dataset and the median with IQR over 10 seed settings is indicated per model. Training was done on the rebalanced dataset. Training was optimized for ICI.

### 3.3 AUC Optimization for Model Training

We also used the AUC as the optimization goal during training. While discrimination is a relevant ability for models in general, it is less relevant for decision making in prognostication tools.

The discriminative performance (i.e., AUC) for AUC-optimized models is shown in **Table 6** when trained with and without ROSE. Again, as with ICI-optimized models, re-balancing did relevantly decrease calibration and had varying small effects on discrimination.

| Model | Calibration (ICI) | | Discrimination (AUC) | |
|---|---|---|---|---|
| | Median | IQR | Median | IQR |
| No re-balancing | | | | |
| *PREDICT v3* | 0.042 | 0.039-0.049 | 0.738 | 0.719-0.770 |
| *f-PREDICT v3* | 0.103 | 0.057-0.127 | 0.803 | 0.803-0.844 |
| RSF | 0.008 | 0.003-0.011 | 0.757 | 0.732-0.797 |
| XGB | 0.044 | 0.039-0.047 | 0.779 | 0.747-0.820 |
| Ensemble | 0.043 | 0.037-0.045 | 0.795 | 0.787-0.813 |
| Re-balancing | | | | |
| *PREDICT v3* | 0.042 | 0.039-0.049 | 0.738 | 0.719-0.770 |
| *f-PREDICT v3* | 0.172 | 0.143-0.242 | 0.828 | 0.797-0.842 |
| RSF | 0.274 | 0.262-0.322 | 0.795 | 0.775-0.810 |
| XGB | 0.563 | 0.551-0.568 | 0.799 | 0.766-0.811 |
| Ensemble | 0.469 | 0.434-0.496 | 0.800 | 0.771-0.813 |

**Table 6. Discrimination (AUC) and Calibration (ICI) When Optimizing for AUC.** AUC and ICI values were calculated on the validation dataset and the median with IQR over 10 seed settings is indicated per model. Training was done on the non-rebalanced or rebalanced dataset. Training was optimized for AUC.